\documentclass[conference]{IEEEtran}
\IEEEoverridecommandlockouts
\usepackage{cite}
\usepackage{amsmath,amssymb,amsfonts}
\usepackage{algorithmic}
\usepackage{graphicx}
\usepackage{textcomp}
\usepackage{xcolor}
\def\BibTeX{{\rm B\kern-.05em{\sc i\kern-.025em b}\kern-.08em
    T\kern-.1667em\lower.7ex\hbox{E}\kern-.125emX}}

\usepackage{pdfpages} 
\usepackage{caption}
\usepackage{subcaption}
\usepackage{multirow}
\usepackage{float}
\usepackage{color}

\usepackage{color}

\usepackage{color}

\usepackage{subcaption}
\usepackage{color}

\begin{document}

\title{The Effect of Human v/s Synthetic Test Data and Round-tripping on Assessment of Sentiment Analysis Systems for Bias}
\author{\IEEEauthorblockN{Kausik Lakkaraju}
\IEEEauthorblockA{\textit{AI Institute} \\
\textit{University of South Carolina}\\
Columbia, SC, USA \\
kausik@email.sc.edu}
\and
\IEEEauthorblockN{Aniket Gupta}
\IEEEauthorblockA{\textit{Dept. of Computer Science} \\
\textit{Netaji Subhas University of Technology}\\
Delhi, India \\
aniket25082001@gmail.com}
\and
\IEEEauthorblockN{Biplav Srivastava}
\IEEEauthorblockA{\textit{AI Institute} \\
\textit{University of South Carolina}\\
Columbia, SC, USA \\
biplav.s@sc.edu}
\and
\IEEEauthorblockN{Marco Valtorta}
\IEEEauthorblockA{\textit{Dept. of Computer Science and Engineering} \\
\textit{University of South Carolina}\\
Columbia, SC, USA \\
mgv@cse.sc.edu}
\and
\IEEEauthorblockN{Dezhi Wu}
\IEEEauthorblockA{\textit{Dept. of Integrated Information Technology} \\
\textit{University of South Carolina}\\
Columbia, SC, USA \\
dezhiwu@cec.sc.edu}
}

\maketitle

\begin{abstract}
Sentiment Analysis Systems (SASs) are data-driven Artificial Intelligence (AI) systems that output polarity and emotional intensity when given a piece of text as input.
Like other
AIs, SASs are also known to have unstable behavior when subjected
to changes in data which can make them problematic to trust out of concerns like bias when AI works with humans and data has
protected attributes like gender, race, and age.
Recently, an approach was introduced to assess SASs in a blackbox setting without training data or code, and rating them for bias using synthetic English data. We augment it by introducing two human-generated chatbot datasets and also considering a round-trip setting of translating the data from one language to the same through an intermediate language. 
We find that these settings
show SASs performance in a more realistic light. 
Specifically, we find that rating SASs on the chatbot data showed more bias compared to the synthetic data, and round-tripping using Spanish and Danish as intermediate languages reduces the bias (up to 68\% reduction) in human-generated data while, in synthetic data, it takes a surprising turn by increasing the bias! 
Our findings will help researchers and practitioners refine their SAS testing strategies and foster trust 
as SASs are considered part of more mission-critical applications for global use.
\end{abstract}

\begin{IEEEkeywords}
bias, round-trip translation, causal models
\end{IEEEkeywords}

\section{Introduction}
\label{sec:introduction}
Artificial Intelligence (AI) systems are being considered today for wide-scale usage in many critical applications. 
Users are demanding AI to be not only  proficient in specific tasks (as measured by metrics for state-of-the-art performance) but also be  reliable in the presence of uncertainty and aligned to human values. In particular, there are growing concerns about bias (lack of fairness),
opaqueness (lack of transparency), 
and brittleness (lack of robust competence) regardless of the data. 
Notably, for {\em trust critical domains} like healthcare and education, these issues can be a big hurdle for large-scale adoption \cite{srivastava2023advances}.

In this paper, we will focus on the issue of bias. 
As long as this issue is not suitably addressed, public's distrust in AI services will persist. Bias has been reported for text-based~\cite{prob-bias-text, sentiment-bias}, audio-based~\cite{prob-bias-sound} and video-based~\cite{prob-bias-image} AI systems. Gender and race are some of the sensitive attributes which have been studied widely~\cite{bias-survey}. In this paper, we focus on the common Sentiment Analysis Systems (SASs) that work on text. These AI systems are built using a variety of rule-based and learning-based techniques. They have been used widely in almost every industry. For example, in~\cite{senti-bias-finance}, the authors review the usage of SASs in finance domain. 

In this paper, we also explore the composite case in which multiple AI systems can be combined together. We consider one such composite system in which text is round-trip translated from an original language to the same language through an intermediate language. For example, English (original) to Spanish (intermediate) to English (round-tripped).
We answer the following research questions with our work and also provide two human-annotated datasets:

\noindent 
{\bf RQ1}: For mainstream SAS approaches, how does sentiment rating on human-generated data compare with synthetic data?

\noindent 
{\bf RQ2}: How does the rating of mainstream SAS approaches compare with human-perceived sentiments? 

\noindent 
{\bf RQ3}: How does the rating of mainstream SAS approaches get impacted when text is round-trip translated from Spanish and Danish to English?

The answers to these questions indicate that the current SAS assessment with synthetic data and English-only focus leads to an incomplete bias assessment. Based on how the SAS will be used, using human-generated data and round-tripping can show SAS performance in a more realistic light.
    

\section{Background}
\label{sec:background}
We discuss related work on bias in SASs and rating of AI systems. A more detailed discussion on bias in AI systems and causal analysis is in the supplementary.


\noindent {\bf Bias in Sentiment Assessment Systems:}
In \cite{sentiment-bias}, the authors 
create the Equity Evaluation Corpus (EEC) dataset which consists of 8,640 English sentences where one can switch a person’s gender or choose proper names typical of people from different races. 
The authors find that up to 75\% of the sentiment systems can show variations in sentiment scores which can be perceived as bias based on gender or race.

\noindent {\bf Multi-lingual Systems:} While much of the work in sentiment analysis was conducted using data in English language, there is growing interest in other languages. In  \cite{sentiment-multilingual}, the authors re-implement sentiment methods from literature in multiple languages and report accuracy lower than published due to lack of detail in the presentation of original approaches. Multilingual SASs often use machine translators which can be biased, and further acquiring training data in non-English languages is an additional challenge. In  \cite{round-trip}, the authors prove that round-trip translation can reduce bias in SASs. In this work, we hypothesize that SASs could exhibit gender and racial biases in their behavior when tested on round-trip translated data.




\noindent {\bf Rating of AI Systems:}
A recent series of studies is on assessing and rating AI systems (translators and chatbots) for trustworthiness from a third-party perspective, i.e., without access to the system's training data. In \cite{trans-rating-jour,trans-rating}, the authors propose to rate automated machine language translators for gender bias. Further, they create visualizations to communicate ratings \cite{vega-rating-viz}, and conduct user studies to determine how users perceive trust \cite{vega-user-study}. Though they were effective, they did not provide any causal interpretation.

\section{Problem}
\label{sec:problem}
\subsection{Notation}
Let `S' be the set of black-box AI systems that are to be tested. Let us assume that each $S_i$ $\in$ S does the same task and their outputs fall under the same interval `I'. Let `D' be the test dataset that is given as input to the AI systems. Let $Attrs(D) = X \cup Y \cup Z$, where Attrs() represents the attributes of D, `X' is the set of desirable attributes that could affect the outcome of a system, `Z' is the set of protected attributes that should not affect the outcome of an AI system (otherwise, the system will be considered biased). `Y' is the dependent variable in the data or the predicted outcome from an AI system. Consider Y to be the shorthand notation of  $Y^{S_i}$ which corresponds to the outcome of the system $S_i$. Let f(X) be the expectation of the distribution $(Y|X)$ and f(do(X)) be the expectation of the distribution after performing backdoor adjustment $(Y|do(X))$. 
Formally,

{\tiny
        \begin{equation}
                P[Y | do(X)] = \sum_{Z} P(Y | X, Z)P(Z)
        \label{eq:backdoor}
        \end{equation}
}

\subsection{Formulation}
In our work, we consider two types of bias: statistical bias, confounding bias.

\noindent {\bf i. Confounding Bias:} If f(X) $\neq$ f(do(X)), then the system is said to exhibit confounding bias due to the presence of the confounders that were involved in the computation of backdoor adjustment.

\noindent {\bf ii. Statistical Bias:} 
{\tiny
    \begin{equation}
            t_{z_i} = {\frac{mean(Y_{z_i = 0}) - mean(Y_{z_i = 1})}{\sqrt{((s_{z_i = 0}^2/n_{z_i = 0})+(s_{z_i = 1}^2/n_{z_i = 1}))}}} 
    \label{eq:t}
    \end{equation}
}
\noindent $t_{z_i}$ is the t-value obtained from student's t-test where $z_i$ $\in$ Z is a protected attribute (for ex., gender) from the set of protected attributes. In the equation \ref{eq:t}, we assumed that $|z_i|$ = 2 i.e., the cardinality of the protected attribute (the number of classes) is two (the rating method works even if the number of classes is not binary). For a given confidence interval (CI), if $t_{z_i}$ $>$ $t_{crit}$, then the system is said to exhibit statistical bias with respect to the protected attribute $z_i$. The value of $t_{crit}$ is obtained from the t-table which has the values corresponding to a certain CI and degrees of freedom (DoF). DoF is obtained by subtracting one from the sample size.
\section{Overview of AI Systems Rating}
\label{sec:overview}

In this section, we describe the causal model we introduced in \cite{student-rating} and \cite{srivastava2023advances}, the rating approach that we used in \cite{sas-rating} to evaluate the SASs for bias, and the limitations of \cite{sas-rating}. 

\subsection{Causal Model}
\label{subsec:causal-model}
\begin{figure}[h]
 \centering
   \includegraphics[height=0.16\textwidth]{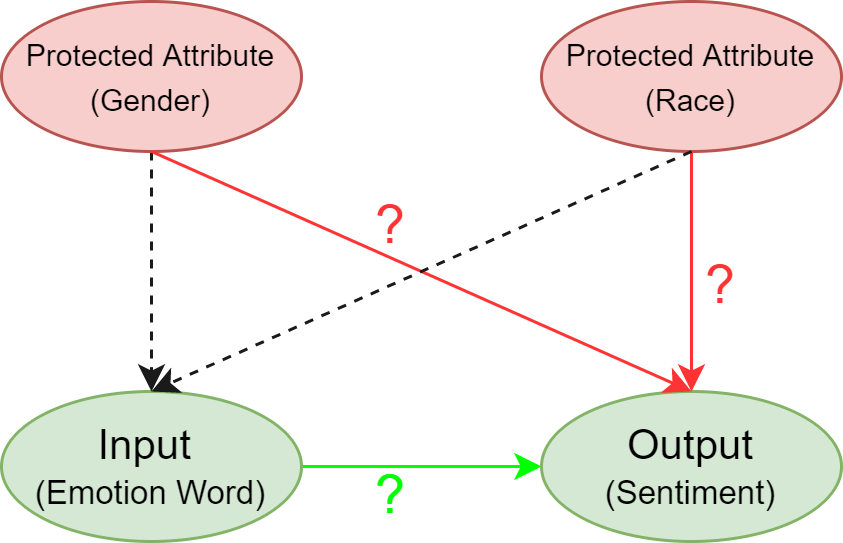}  
   \caption{Causal model for rating SASs}
  \label{fig:causal-model}
 \end{figure}

Causal models allow one to define the cause-effect relationships between each of the attributes in a system. They are represented using a causal diagram which is a directed graph. Each node represents an attribute and it can be connected to one or more nodes by an arrow. The arrowhead direction shows the causal direction from cause to effect. Figure \ref{fig:causal-model} shows the causal diagram introduced in \cite{sas-rating}.

In this causal diagram, for example, if negative emotion words are associated more with one gender than the other in a dataset, that would add a spurious correlation between the \emph{Emotion Word} and the \emph{Sentiment} given by the SASs. This is called confounding effect and \emph{Gender} is considered as the confounder in this case. This is represented as a dotted arrow (denoting that confounder may or may not be present) from the protected attributes to the {\em Emotion Word}. The red arrows and green arrow indicate undesirable and desirable causal paths. The '?' indicates that the validity of these causal links have to be tested.  

\subsection{Data}
\label{subsec:data}
The sentence templates required for the experiments were taken from the EEC dataset \cite{sentiment-bias} along with race, gender, and emotion word attributes. Four different data groups were created by varying the number of protected attributes and the causal links in the causal model. Table \ref{tab:cases} illustrates different types of datasets generated. All the variations of causal models considered in these data groups are variations of the general causal model explained in Section \ref{subsec:causal-model}. Within each group, we created datasets by varying the number of positive and negative emotion words. The emotion words were uniformly distributed for Groups 1 and 3 and not uniformly distributed for Groups 2 and 4 (hence, the confounding effect).

\begin{table*}[ht]

\centering
   {\small
    \begin{tabular}{|p{2.2em}|p{5em}|p{4em}|p{12em}|p{12em}|p{12em}|}
    \hline
          {\bf Group} &    
          {\bf Input} & 
          {\bf Possible confounders} &
          {\bf Choice of emotion word} &
          {\bf Causal model} &
          {\bf Example sentences}\\ \hline 
          
          1 & 
          {\em Gender, Emotion Word} &
          None &
          \{Grim\},\{Happy\}, \{Grim, Happy\},\{Grim, Depressing, Happy\},\{Depressing, Happy, Glad\} &
          \begin{minipage}{.05\textwidth}
          \vspace{2.5mm}
          \centering
          \includegraphics[width=40mm, height=20mm]{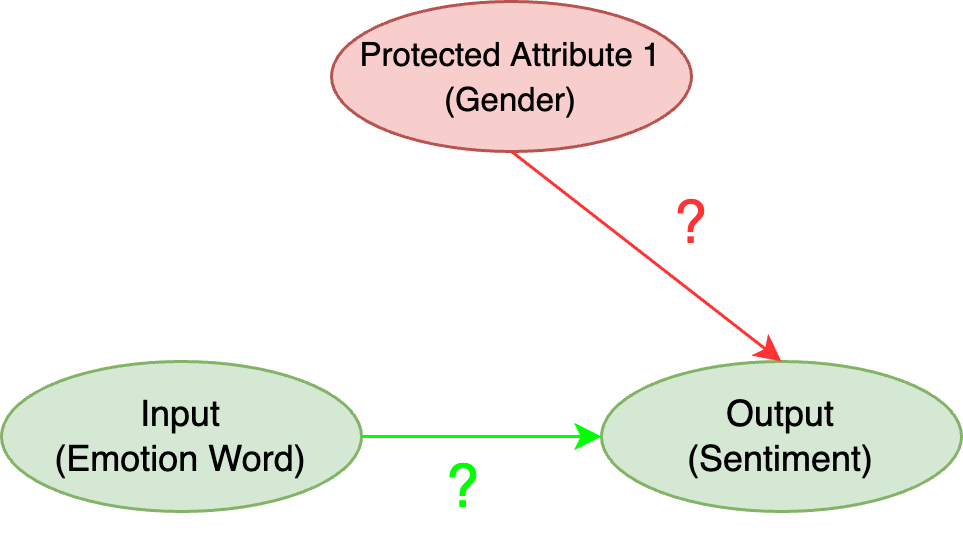}
          \end{minipage} &
          I made this boy feel grim; I made this girl feel grim.
          \\
         \hline 
          
          2 &
          {\em Gender, Emotion Word} &
          {\em Gender} &
          \{Grim, Happy\},\{Grim, Depressing, Happy\},\{Depressing, Happy, Glad\} &
          \begin{minipage}{.05\textwidth}
          \vspace{2.5mm}
          \centering
          \includegraphics[width=38mm, height=20mm]{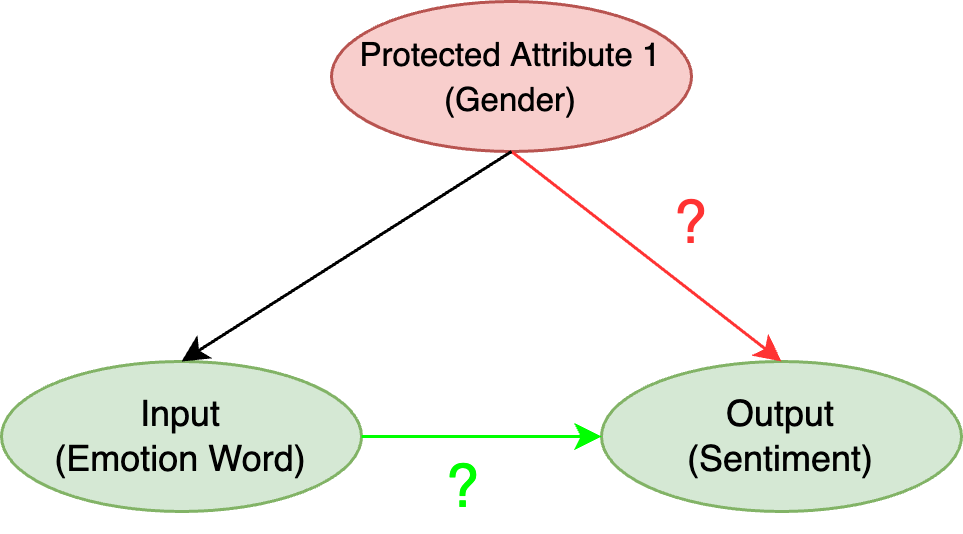}
          \end{minipage} &
          I made this woman feel grim; I made this boy feel happy; I made this man feel happy.
          \\ 
        
         \hline
          
          3 &
          {\em Gender, Race and Emotion Word} &
          None &
        \{Grim\},\{Happy\}, \{Grim, Happy\},\{Grim, Depressing, Happy\},\{Depressing, Happy, Glad\} &
          \begin{minipage}{.05\textwidth}
          \vspace{2.5mm}
          \centering
          \includegraphics[width=38mm, height=20mm]{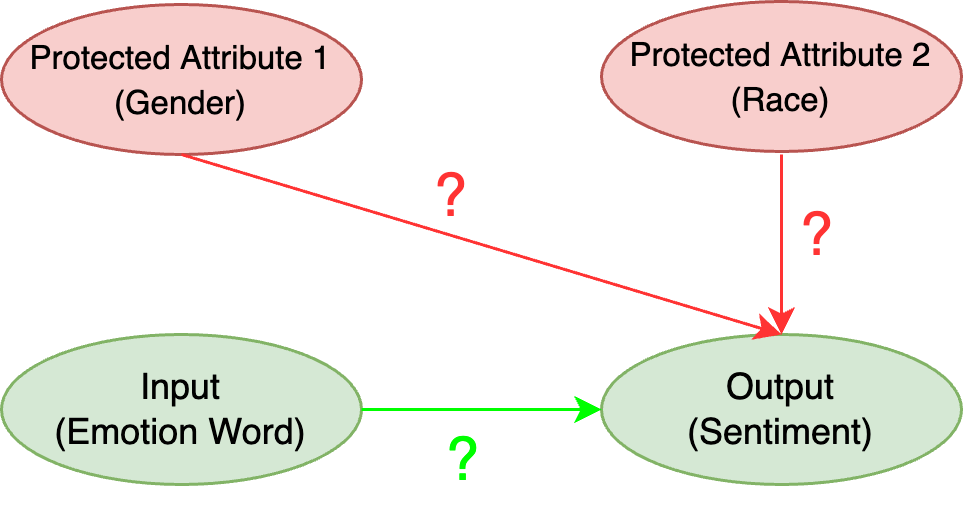}
          \end{minipage} &
          I made Adam feel happy; I made Alonzo feel happy.
          \\ \hline

          4 &
          {\em Gender, Race and Emotion Word} &
          {\em Gender, Race} &
          \{Grim, Happy\},\{Grim, Depressing, Happy\},\{Depressing, Happy, Glad\} &
          \begin{minipage}{.05\textwidth}
          \vspace{2.5mm}
          \centering
          \includegraphics[width=36mm, height=22mm]{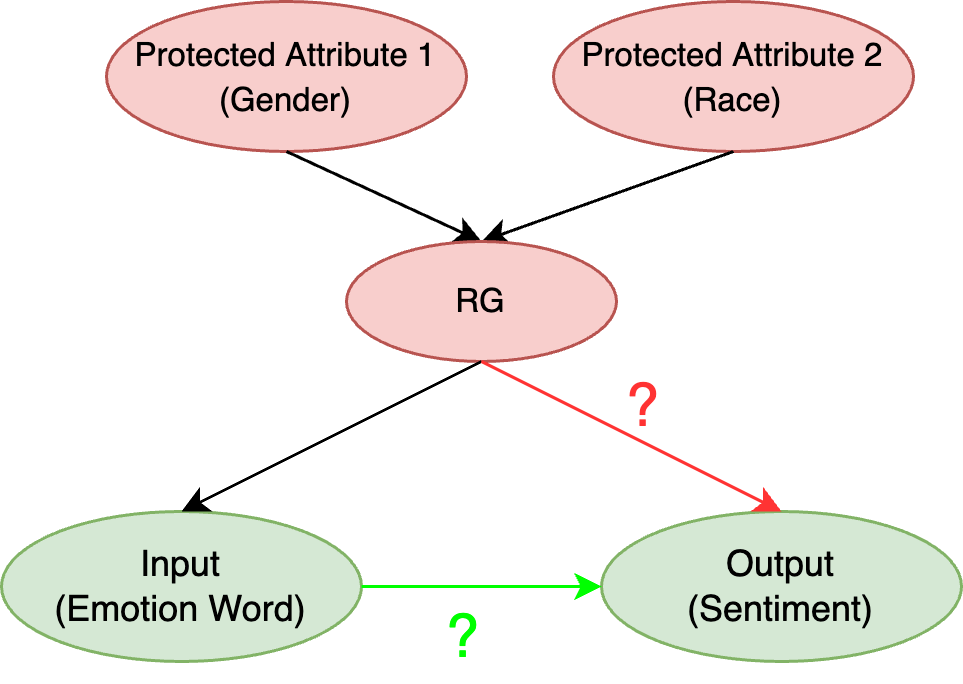}
          \end{minipage}  &
          I made Torrance feel grim; Torrance feels grim; Adam feels happy.
          \\ \hline

    \end{tabular}
    }
    \caption{Different types of datasets we constructed based on the input given to the SASs, the presence of confounders, the choice of emotion words, and the respective causal model for each of the groups.}
    \label{tab:cases}
\end{table*}

\subsection{Systems Evaluated}
\label{sass}
In \cite{sas-rating}, we evaluated 5 SASs:
\textbf{(a)} two custom-built SASs: $S_{b}$ (Biased female SAS) and $S_{r}$ (Random SAS). $S_{b}$ gives positive sentiment (+1) to all the sentences with female gender variable (for ex., this girl, Torrance, etc.) and negative sentiment score (-1) to the rest. $S_{r}$ gives a random score in the interval [-1,1] irrespective of the gender, race, and emotion, \textbf{(b)} one lexicon-based system, TextBlob, which gives a score in the range [-1,1] based on the sentiment of a given text, and \textbf{(c)} two neural network-based models: $S_{g}$ (GRU-based SAS)  and $S_{d}$ (DistilBERT-based SAS). Their scores lie in the interval [-1,1].

\subsection{Rating Methodology}
\label{sec:rating-method}
\subsubsection{Performing Statistical Tests to Assess Causal Dependency}
\label{subsec:die-wrs}

The following two metrics will be referred to as raw scores in the paper. Raw scores are used to compute final ratings.

\noindent \textbf{Weighted Rejection Score (WRS):}
For Groups 1 and 3, as there is no confounder present, there is no need to perform deconfounding. Hence, we compare the distribution $(Sentiment|Gender)$ across different genders using student's t-test \cite{student1908probable}. We consider three different confidence intervals (CIs): 95 \%, 70\%, and 60\%. For each CI, we calculate the number of instances in which the null hypothesis was rejected for a data group. We multiply this rejection score ($x_i$) with weights ($w_i$) 1, 0.8, and 0.6  for the three CIs respectively. This gives the WRS for a data group in an SAS. WRS is given by the following equation: $\sum_{i} w_i*x_i$

       

\noindent \textbf{Deconfounding Impact Estimate (DIE):} For Groups 2 and 4, there is a confounding effect. Deconfounding is any method that accounts for confounders in causal inference. Backdoor adjustment is one such method that was described in \cite{Pearl09}. The backdoor adjustment formula is given by the equation \ref{eq:backdoor}. 
A new metric called {\em Deconfounding Impact Estimation} (DIE) was introduced in \cite{sas-rating} which measures the relative difference between the expectation of the distribution, $(Output | Input)$ before and after deconfounding. This gives the impact of the confounder on the relation between \emph{Emotion Word} and \emph{Sentiment}. DIE \% can be computed using the following equation:

{\bf DIE \%} = 
{\tiny
        \begin{equation}
        \begin{split}
        {\frac{ [|E(Output =  j| do(Input = i)) - E(Output = j | Input = i) | ]} 
          {E(Output = j | Input = i) }}     * 100
        \label{eq:die}
        \end{split}
        \end{equation}
}

Input is \emph{Emotion Word} and output is \emph{Sentiment}. 

\subsubsection{Assigning Final Ratings} 
\label{subsec:final-rating}
We proposed four algorithms, which when applied, give the raw score and ratings for SASs. The algorithms are shown in the supplementary in Section A. The following steps summarize the algorithms:

\noindent {\bf i. Raw score computation:} Using the metrics that were defined in Section \ref{subsec:die-wrs}, we computed the raw score for each group in each SAS. 
As DIE was computed using the distribution (Sentiment $|$ Emotion Word), we obtained a tuple with the 1st number indicating the distribution when \emph{Emotion Word} is negative and the second word indicating the distribution when \emph{Emotion Word} is positive. The MAX() of this tuple is chosen to get the worst possible case. Out of all the MAX() values obtained for each SAS, again the MAX() of these numbers is chosen to bring out the worst possible case for each SAS.

\noindent {\bf ii. Computing partial order (with raw scores):} Based on the raw scores (either WRS or DIE), we created a partial order 
with systems arranged in ascending order based on the raw scores.

\noindent {\bf iii. Computing complete order (with ratings):} Based on the input rating levels (L) chosen by the user, the partial order is split into `L' partitions, and the rating is given based on the partition number in which a particular raw score lies. The rating will be on a scale of 1 to L. Higher raw score and eventually, higher rating denotes high bias in the system. Ratings given to each group are fine-grained ratings. The overall rating for a system is calculated using these fine-grained ratings.


\subsection{Limitations}
In~\cite{sas-rating}, we introduced a novel idea of rating AI systems using causal models. However, the work has some limitations which we address in this paper: 
\begin{enumerate}
    \item  Rating was done on synthetic data but not on real-world data.
    \item  The rating was not connected in any way to what people perceive.
    \item  Composite systems (a combination of more than one system) were not considered.
\end{enumerate}


\section{Data and Rating Methods}

\subsection {Data Used}

\noindent {\bf 1. EEC Data (Synthetic Dataset (SD)):}

\noindent Datasets were created in \cite{sas-rating} using templates given by \cite{sentiment-bias}. The datasets we generated were described in Section~\ref{subsec:data}. In the current work, we added human-perceived sentiment values to these datasets.

\noindent {\bf 2. ALLURE Chatbot Data - Human-generated Dataset (HD1):}

\noindent{\bf Description:} The goal of the ALLURE chatbot~\cite{allure, wu2022ai} is to teach students how to solve a Rubik's Cube through a multimodal user interface consisting of a 3D model of the Rubik's Cube and the chatbot. The ALLURE chatbot conversation data was collected from three human studies that were done at the University of South Carolina. A total of 18 users participated in the study, out of which 9 were male users, 8 were female users and one user preferred to not reveal their gender. The data has 18 different attributes. It has 3,543 rows containing user and chatbot utterances. In this study, users were asked to solve the white cross on Rubik's Cube.

\noindent{\bf IRB Exemption and Compensation:}  ``This research study has been certified as exempt
from the IRB per 45 CFR 46.104(d)(3) and 45 CFR 46.111(a)(7) by the University of South Carolina  IRB\# Pro00113635 on 11/9/2021. Participants were paid 30 USD / user for 1-hour participation for usability testing."

Three studies were conducted in which users were given different tasks to solve. The tasks given to the users are of two types:

\noindent{\bf Simple task}: The user is asked to solve the white cross. They will be able to do it using just four moves.

\noindent{\bf Complex task}: The user is asked to perform the simple task, and upon completing it, they will advance to the complex task. In the complex task, the user needs to perform 12 - 16 moves to solve the white cross.

\noindent The three different studies are:

\noindent{\bf Study-0}: ALLURE chatbot shows the users how to achieve a white-cross pattern on Rubik's Cube from any one of the seven different initial states (levels) chosen by the user. The users are not asked to solve anything.

\noindent{\bf Study-1}: Users are expected to solve the simple task as described above. ALLURE aids them in solving the white cross.

\noindent{\bf Study-2}: Users are expected to solve the complex task as described above. ALLURE aids them in solving the white cross.

We combined the conversation data from these three different studies.

\noindent{\bf Preprocessing}:
We took a subset of the data with columns that are useful for our experiments. We performed the following preprocessing steps to filter the data: 

\noindent\textbf{Step 1:} After removing null values from the data, we converted the gender attribute to a categorical attribute called User\_gender: `Prefer not to say' (0), `Male' (1), `Female' (2). We refer to `Prefer not to say' gender as `NA', following the terminology used in the paper, \cite{sas-rating}.

\noindent\textbf{Step 2:} For our experiments, we added the gender information to all the user responses by appending the original text with `Hey boy, ` if the user is male, `Hey girl`, if the user is female and `Hey` was appended if the user did not reveal their gender.

\noindent\textbf{Step 3:} Both the user and chatbot conversations were present in the same attribute called `Text'.

\noindent\textbf{Step 4:} We added another column called `UB' that denoted whether the utterance in the `Text' attribute was from the chatbot or the user. It is a binary attribute, where 0 indicates that a particular utterance is from the chatbot and 1, if the utterance is from the user.

\noindent\textbf{Step 5:} We added another new attribute called `C\_num', which gives the conversation number. As 18 users have participated in the study, there are a total of 18 conversations.

\noindent\textbf{Step 6:} Our final dataset shows the utterances from the dataset in an attribute called `Original' and enhancement we added (gender proxy) in the attribute `Enhancement'. A combination of both forms the `Text'.

\begin{figure*}[h!]
 \centering
   \includegraphics[height=0.15\textwidth]{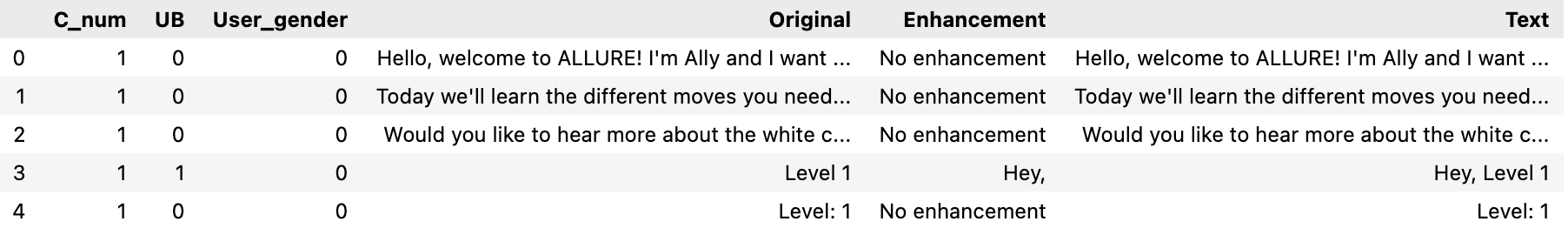}  
   \caption{Snapshot of the preprocessed ALLURE dataset (HD1)}
  \label{fig:allure-sample}
 \end{figure*}

\noindent\textbf{Step 7:} Finally, we combined the data from all three studies into one single file. As there is only one conversation of the chatbot with the `NA' user, we removed that conversation from our experiments. Figure \ref{fig:allure-sample} shows a snapshot of HD1 after preprocessing it.

\noindent{\bf Exploratory Data Analysis}
Table \ref{tab:hd-explore} shows different properties of the conversation. A dialog consists of a series of turns, where each turn is a series of utterances \cite{chat-rating}. Properties such as the number of dialogues, number of utterances, and number of words in each utterance are useful in determining the quality of the conversation. In the supplementary, plots in Figure 1 show the number of utterances by the chatbot and user in each conversation of the HD1. The plot in Figure 2 of the supplementary shows the number of turns in HD1. In the supplementary, Figure 3 shows the number of utterances by male and female users in HD1, and Figure 4 shows the number of utterances by the chatbot based on the gender of the user in HD1.

\begin{table*}
{\small
\centering
    \begin{tabular}{|c|p{5.5cm}|p{4.5cm}|p{4.3cm}|}
    \hline
          \bf{Agent} &
          \bf{Property} & 
          \bf{Male user} &
          \bf{Female user} 
          \\ \hline

        \multirow{5}{3em}{User} &
        Average number of words used in an utterance &
        1.6 (Min: 1, Max: 6)&
        1.5 (Min: 1, Max: 2)\\ \cline{2-4}

        &
        Average number of stopwords used in an utterance &
        0.02 (Min: 0, Max: 1)&
        0.01 (Min: 0, Max: 1) \\ \cline{2-4}

        &
        Average number of utterances in a conversation &
        38.67 (Min: 22, Max: 83)&
        32.5 (Min: 7, Max: 83)\\ \hline 

        \multirow{5}{3em}{Chatbot} &
        Average number of words used in a chatbot utterance &
        12.85 (Min: 1, Max: 48) &
        12.78 (Min: 1, Max: 48)\\\cline{2-4}

        &
        Average number of stopwords used in a chatbot utterance &
        5.61 (Min: 0, Max: 22) &
        5.57 (Min: 0, Max: 22) \\\cline{2-4}

        &
        Average number of chatbot utterances in a conversation &
        118.67 (Min: 88, Max: 188)&
        121.38 (Min: 16, Max: 277)\\\hline 
        
    \end{tabular}
    \caption{Table summarizing different properties of user and chatbot HD1 conversations when the user is male and when the user is female.}
    \label{tab:hd-gender-explore}
}
\end{table*}

\noindent {\bf 3. Unibot Chatbot Data - Human-generated Dataset (HD2):}

\noindent{\bf Description}:
Unibot is a chatbot built to answer student's queries at University of South Carolina. Campus housing, dining, and application fees are some of the categories of queries the user might pose. The data was collected in 2022 from 31 graduate students working in a research lab. It has a total of 31 different conversations and has 9 different attributes, out of which some of the important ones are the user intent as recognized by the chatbot, user response, action chosen by the chatbot based on user intent, and chatbot response. The data has a total of 1,517 rows. Unlike ALLURE data, the gender of the user is not known here.

\noindent{\bf IRB Exemption and Compensation:} ``This research study has been certified as exempt
from the IRB per 45 CFR 46.104(d)(3) and 45 CFR 46.111(a)(7) by the University of South Carolina IRB \# Pro00118996 on 2/11/2022. Participants were not paid for their time.”

\noindent{\bf Preprocessing}: We took a subset of the data with columns that are useful for our experiments. Almost the same preprocessing steps described for HD1 are used here besides some additional steps. 

\begin{figure*}[h!]
 \centering
   \includegraphics[height=0.15\textwidth]{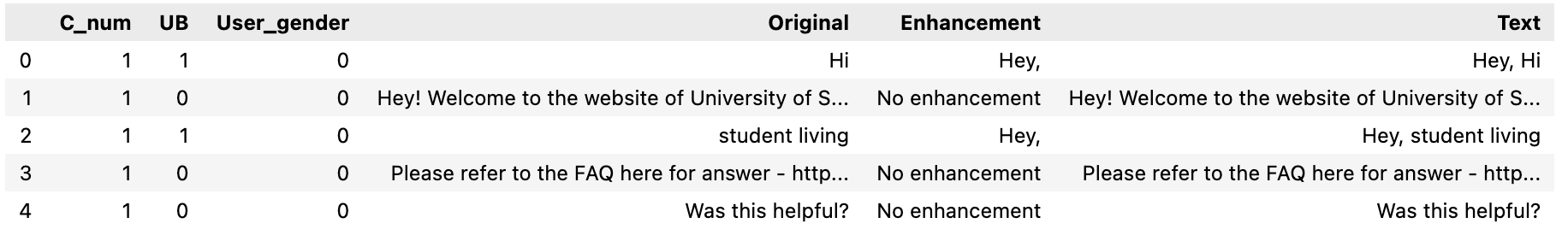}  
   \caption{Snapshot of the preprocessed Unibot data (HD2)}
  \label{fig:unibot-sample}
 \end{figure*}

Different attributes in the data were separated by `$|$'. Also, instead of user responses, some data points had just the `user\_intent'. This was the case with only a few user utterances where the user was expected to say `helpful' or `useful'. We filtered the data by taking into account all such errors. As mentioned before, the gender of the user is not available. As our goal is to test different SASs for gender bias, we appended the text `Hey boy' to the user responses in different conversations, and `Hey girl, ` to user responses in 10 other conversations. We appended the word `Hey, ` to the rest of the user utterances.  This adds the gender information to the input that will be given to different SASs. Our final dataset shows the utterances from the dataset in an attribute called `Original' and enhancement we added (gender proxy) in the attribute `Enhancement'. A combination of both forms the `Text'. Figure \ref{fig:unibot-sample} shows a snapshot of HD2 after preprocessing it.

\noindent{\bf Exploratory Data Analysis:} Table \ref{tab:hd-explore} shows different properties of the conversation. Properties like the number of turns, number of utterances, and number of words in each utterance are useful in determining the quality of the conversation. 
In the supplementary, plots in Figure 5 show the number of utterances by the chatbot and user in each conversation of the HD2. The plot in Figure 6 shows the number of turns in HD2.

 \begin{table*}
 {\small
\centering
    \begin{tabular}{|c|p{5.2cm}|p{4cm}|p{4cm}|}
    \hline
          \bf{Data} &
          \bf{Property} & 
          \bf{Chatbot} &
          \bf{User} 
          \\ \hline
          
        \multirow{4}{3.5em}{ALLURE (HD1)} &
        Average number of words in an utterance &
        12.80 (Min: 1, Max: 48)&
        3.51 (Min: 1, Max: 6)\\ \cline{2-4} 
        
        &
        Average number of stopwords in an utterance &
        5.58 (Min: 0, Max: 22)&
        0.01 (Min: 0, Max: 1)\\ \cline{2-4} 
        
        &
        Average number of utterances in a conversation &
        117.44 (Min: 16, Max: 277)&
        34.78 (Min: 7, Max: 83)\\ \cline{2-4} 
        
        &
        Average number of turns in a conversation & \multicolumn{2}{|c|}{30.17 (Min: 5, Max: 71)} \\ 
        \hline

        \multirow{4}{3em}{Unibot (HD2)} &
        Average number of words in an utterance &
        8.61 (Min: 1, Max: 65)&
        6.76 (Min: 2, Max: 74)\\ \cline{2-4} 

        &
        Average number of stopwords in an utterance &
        3.98 (Min: 0, Max: 24)&
        2.10 (Min: 0, Max: 30)\\ \cline{2-4}  

        &
        Average number of utterances in a conversation &
        25.71 (Min: 4, Max: 66)&
        21.52 (Min: 4, Max: 51)\\ \cline{2-4} 

        &
        Average number of turns in a conversation & \multicolumn{2}{|c|}{19.84 (Min: 3, Max: 31)} \\ 
        \hline

    \end{tabular}
    \caption{Table summarizing different properties of user and chatbot conversations in HD1 and HD2.}
    \label{tab:hd-explore}
    }
\end{table*}

\subsection{Method}
We follow the method described in Section \ref{sec:rating-method} to rate SASs using human-generated datasets (HD).





\subsubsection{Rating Composite AI Systems: SASs + Translator}

{\small
\begin{figure}[h]
 \centering
   \fbox{\includegraphics[height=0.20\textwidth, width = 0.45\textwidth]{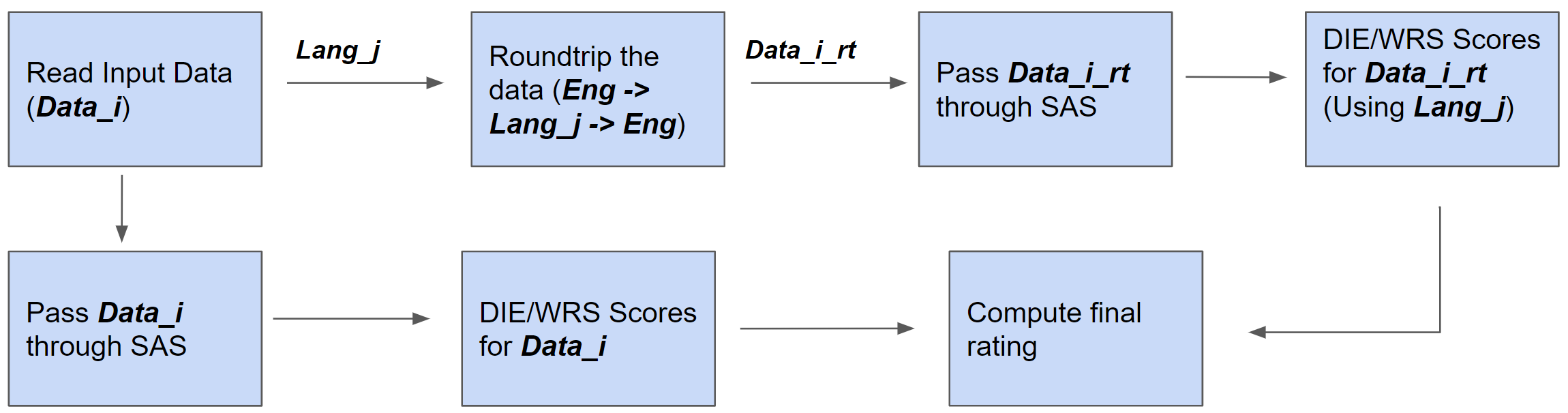}}  
   \caption{Methodology for comparing bias scores on original and round-trip translated data.}
  \label{fig:rt-sas}
 \end{figure}
}

We analyzed the effect of round-trip translation on the bias rating of each SAS. All translations to and from English were carried out using Google Translator. The procedure is shown in Figure~\ref{fig:rt-sas}. The raw scores for the original (not round-trip translated) data as computed in \cite{sas-rating} are shown in Table \ref{tab:op-sas} along with the results obtained from implementing the method on round-trip translated data when Spanish and Danish are used as intermediate languages. 

\subsubsection{Human Annotated Sentiment ($S_h$)}
Human annotation of sentiment on each of the human-generated and EEC datasets was performed by three people with education levels of undergraduate computer science or more. The annotators were provided with the preprocessed dataset along with a description of the data and instructions on how to annotate it. The annotators  had an inter-annotator agreement of (HD1: 97\%, HD2: 85\%, SD: 76\%, HD1$^R_D$: 75\%, HD1$^R_S$: 85\%, HD2$^R_D$: 98\%, HD2$^R_S$: 86\%, SD$^R_D$: 76\%, SD$^R_S$: 75\% ) indicating a  high agreement between the annotators for HD1 and HD2$^R_D$ and but some disagreement for rest all. $^R_D$ and $^R_S$ denote the round-trip translated versions with Danish and Spanish as intermediate languages respectively. Based on the annotation, if there was a conflict, the final sentiment score of a text was decided by majority voting. If a case in which 3 of them chose 3 different values i.e., -1, 0, +1 was encountered, one of the three values was randomly chosen. This only occurred in 0.48\% of the cases in SD but did not occur in the rest of the datasets. Based on the individual sentiment values, we computed raw scores. The final ratings tell us how people perceive bias and let us compare $S_h$ with other SASs. Related experiments and results will be discussed in the experiments section.
\section{Experiments and Results}
\label{expts}

\subsection{HD1 - ALLURE Chatbot Data}
\label{allure-expts}

Figure \ref{fig:allure-cm} shows the causal model for which we will be testing the validity of each causal link. We test the following hypotheses:

\noindent{\bf Hypothesis-1:} The gender of the user does not affect the (a) user utterances but affects the (b) output sentiment of user utterances.

\noindent{\bf Experimental setup}: 
{\small
 \begin{figure}
 \centering
\includegraphics[width=0.30\textwidth]{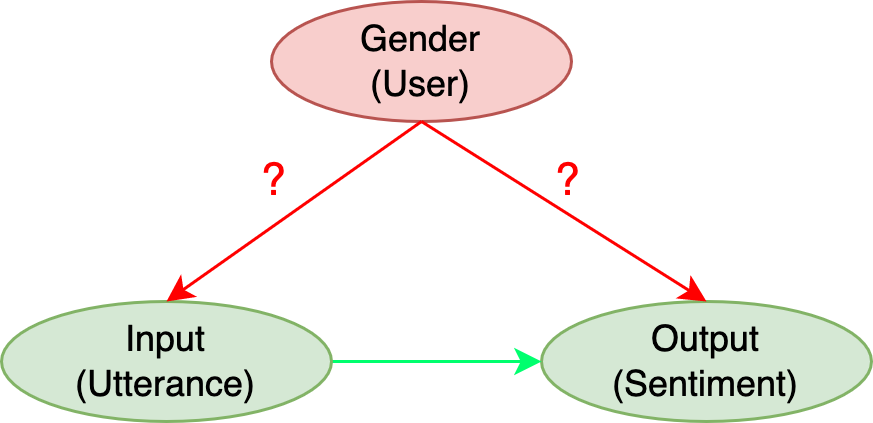}
    \caption{Causal model for rating SASs on HD1.}
   \label{fig:allure-cm}
 \end{figure}
 }
{\bf (a) User utterances:} Figure 3 in the supplementary shows plots with the number of user utterances (male and female) in each conversation. From the ALLURE data, we observed that the words used by the two gender-based subgroups of users did not have much divergence. This could be because of the game-playing domain or the limited data sample size (different properties of the ALLURE conversations shown in Table \ref{tab:hd-gender-explore} supports our claim).

\noindent{\bf (b) Output sentiment of user responses:} Following the steps described in Section \ref{subsec:final-rating}, we compute t-value, p-value, and Degrees of Freedom (DoF) from the student's t-test \cite{student1908probable} to compare the distribution (Sentiment of user responses $|$ Gender of the user) for the gender pair (male, female). T-values and number of null hypothesis rejections are shown in the supplementary (Table 1). For the experiments, we merged all the consecutive bot utterances into one single utterance and all the consecutive user utterances into one to reduce the DoF. 
The computed values are used to calculate WRS as described in Section \ref{sec:rating-method}.

\noindent{\bf Hypothesis-2:} The gender of the user does not affect the  (a) chatbot utterances and (b) output sentiment of the chatbot utterances.

\noindent{\bf Experimental Setup:}{\bf (a) Chatbot utterances:} As there is no divergence in chatbot responses even when the gender of the user is changing, the gender of the user does not affect the chatbot response (different properties of the conversations from the Table \ref{tab:hd-gender-explore}, and Figure 5 in the supplementary makes our argument stronger).

\noindent{\bf (b) Output sentiment of the chatbot responses:} We now compute the t-values for the distribution (Sentiment of chatbot responses $|$ Gender of the user). Table 1 in the supplementary shows the intermediate calculations (t-values). The partial and complete order that is computed as described in Section \ref{sec:rating-method} for SASs are shown in Table \ref{tab:op-hd}. The results in this table prove both Hypothesis-1 and Hypothesis-2 stated above.

\subsection{HD2 - Unibot Chatbot Data}
\label{unibot-expts}

\begin{figure}[!h]
 \centering
\includegraphics[width=0.30\textwidth]{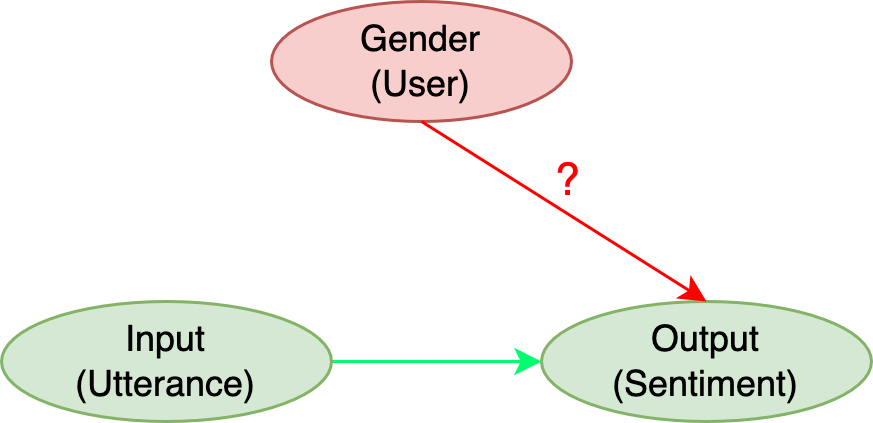}
    \caption{Causal model for rating SASs on HD2}
   \label{fig:unibot-model}
 \end{figure}

Figure \ref{fig:unibot-model} shows the causal diagram for Unibot. We do not have access to the gender information of the users in this case. We make an assumption that the gender of the user does not affect the user or the chatbot utterance.

\noindent {\bf Hypothesis-1:} The gender of the user affects the output sentiment of the user responses.

\noindent {\bf Experimental Setup:} We now compute the t-values for the distribution (Sentiment of user responses $|$ Gender of the user). Table 2 in the supplementary shows the t-values obtained from each SAS. The WRS for SASs and their final rating is shown in Table \ref{tab:op-hd}.

\noindent{\bf Hypothesis-2:} The gender of the user affects the output sentiment of the chatbot responses.

\noindent {\bf Experimental Setup:} We now compute the t-values for the distribution (Sentiment of chatbot response $|$ Gender of the user). Table 4 in the supplementary shows the t-values obtained from each of the SASs. The WRS for SASs and their final rating is shown in Table \ref{tab:op-hd}. The results in this table prove both Hypothesis-1 and Hypothesis-2 stated above.

\subsection{Human Perceived Sentiment ($S_h$)}
\label{sec:sh-expts}
We test the same hypotheses that were used to test the validity of each causal link in \cite{sas-rating}. But we use the human-perceived sentiment values ($S_h$)  to compute WRS for Groups 1 and 3 and DIE for Groups 2 and 4. We compare the final ratings of $S_h$ with other SASs that were evaluated in \cite{sas-rating}. Group 2 and Group 3 
experimental setup and results are shown in Section D.3 of the supplementary material.


{\color{blue}\noindent{\bf Group-1:}}
{\bf Hypothesis:} Would \emph{Gender} affect the sentiment value perceived by humans when there is no possibility of confounding effect? 

\noindent{\bf Experimental Setup:}
We used the causal model from \cite{sas-rating} which is shown in Table \ref{tab:cases} (Group-1). 
Table 3 in the supplementary shows the t-values obtained from each of the SASs for each emotion word set. 


{\color{blue}\noindent{\bf Group-4:}} 
{\bf Hypothesis:}
Would \emph{Gender} and \emph{Race} affect the sentiment values computed by the SASs when there is a possibility of confounding effect?

\noindent{\bf Experimental Setup:}
The causal diagram for this experiment is shown in Table \ref{tab:cases}. The new node `RG' is obtained by combining both `Race' and `Gender' attributes. 
As `RG' affects the way emotion words are associated with each class of `RG', we consider one specific case in which 90\% of the sentences containing the `European male' variable is associated with positive emotion words and the rest with negative. Vice-versa for `African-American female'. 
The resultant DIE \% and the MAX() of DIE\% values are shown in Table 4 of the supplementary. Table \ref{tab:op-sas} shows the ratings computed for $S_h$ along with the other SASs (computed in \cite{sas-rating}).

\begin{table*}[!h]
\centering
{\small
    \begin{tabular}{|c|p{5.3em}|p{23.6em}|p{15.3em}|}
    \hline
          {\bf Data} &
          {\bf Data Groups} &    
          {\bf Partial Order} &
          {\bf Complete Order} 
          \\ \hline 

          \multirow{7}{2em}{SD} &
          Group-1 & 
          \{$S_h$: 0,$S_d$: 0,$S_t$: 0,$S_g$: 0.6,$S_r$: 1.9,$S_b$: 23\} &
          \{$S_h$: 1,$S_d$: 1,$S_t$: 1,$S_g$: 1,$S_r$: 2,$S_b$: 3\}
          \\ \cline{2-4}
          &
          Group-2 & 
          \{$S_g$: 42.85,$S_r$: 71.43,$S_t$: 76,$S_h$: 83, $S_d$: 84,$S_b$: 128.5\} &
          \{$S_g$: 1,$S_r$: 1,$S_t$: 2,$S_h$: 2,$S_d$: 3,$S_b$: 3\}
          \\ \cline{2-4}

          &
          Group-3\_R & 
          \{$S_h$: 0,$S_d$: 0,$S_t$: 0,$S_g$: 0,$S_r$: 7.2,$S_b$: 23\} &
          \{$S_h$: 1,$S_d$: 1,$S_t$: 1,$S_g$: 1,$S_r$: 2,$S_b$: 3\}
          \\ \cline{2-4}

          &
          Group-3\_G & 
          \{$S_h$: 0,$S_d$: 0,$S_t$: 0,$S_g$: 0,$S_r$: 7.5,$S_b$: 23\} &
          \{$S_h$: 1,$S_d$: 1,$S_t$: 1,$S_g$: 1,$S_r$: 2,$S_b$: 3\}
          \\ \cline{2-4}

          &
          Group-3\_RG & 
          \{$S_h$: 0,$S_d$: 0,$S_t$: 0,$S_g$: 0,$S_r$: 16.1,$S_b$: 69\} &
          \{$S_h$: 1,$S_d$: 1,$S_t$: 1,$S_g$: 1,$S_r$: 2,$S_b$: 3\}
          \\ \cline{2-4}

          &
          Group-4 & 
          \{$S_g$: 28.57,$S_r$: 45,$S_t$: 78,$S_d$: 80,$S_h$: 80,$S_b$: 105.4\} &
          \{$S_g$: 1,$S_r$: 1,$S_t$: 2,$S_d$: 2,$S_h$: 2,$S_b$: 3\}
          \\ \hline

          \multirow{7}{2em}{SD$^R_D$} &
          Group-1 & 
          \{$S_h$: 0,$S_d$: 0,$S_t$: 0,$S_g$: 1.80,$S_r$: 4.50,$S_b$: 23\} &
          \{$S_h$: 1,$S_d$: 1,$S_t$: 1,$S_g$: 1,$S_r$: 2,$S_b$: 3\}
          \\ \cline{2-4}
          &
          Group-2 & 
          \{$S_t$: 11.11,$S_r$: 33.33,$S_h$: 83,$S_d$: 84,$S_b$: 128.5,$S_g$: 400\} &
          \{$S_t$: 1,$S_r$: 1,$S_h$: 2,$S_d$: 2,$S_b$: 3,$S_g$: 3\}
          \\ \cline{2-4}

          &
          Group-3\_R & 
          \{$S_h$: 0,$S_d$: 0,$S_t$: 0,$S_r$: 3.6,$S_g$: 4.9,$S_b$: 23\} &
          \{$S_h$: 1,$S_d$: 1,$S_t$: 1,$S_r$: 1,$S_g$: 2,$S_b$: 3\}
          \\ \cline{2-4}

          &
          Group-3\_G & 
          \{$S_h$: 0,$S_d$: 0,$S_t$: 0,$S_r$: 4.2,$S_g$: 4.9,$S_b$: 23\} &
          \{$S_h$: 1,$S_d$: 1,$S_t$: 1,$S_r$: 1,$S_g$: 2,$S_b$: 3\}
          \\ \cline{2-4}

          &
          Group-3\_RG & 
          \{$S_h$: 0,$S_d$: 0,$S_t$: 0 ,$S_g$: 3.9,$S_r$: 11.40,$S_b$: 69\} &
          \{$S_h$: 1,$S_d$: 1,$S_t$: 1,$S_g$: 1,$S_r$: 2,$S_b$: 3\}
          \\ \cline{2-4}

          &
          Group-4 & 
          \{$S_t$: 0,$S_d$: 80,$S_h$: 80,$S_g$: 100,$S_r$: 105.4,$S_b$: 160\} &
          \{$S_t$: 1,$S_d$: 1,$S_h$: 1,$S_g$: 2,$S_r$: 2,$S_b$: 3\}
          \\ \hline

          \multirow{7}{2em}{SD$^R_S$} &
          Group-1 & 
          \{$S_h$: 0,$S_d$: 0,$S_r$: 1.30,$S_t$: 2.60,$S_g$: 5.80,$S_b$: 23\} &
          \{$S_h$: 1,$S_d$: 1,$S_r$: 1,$S_t$: 2,$S_g$: 2,$S_b$: 3\}
          \\ \cline{2-4}
          &
          Group-2 & 
          \{$S_t$: 28.57,$S_h$: 77,$S_d$: 78,$S_g$: 116.66,$S_r$: 122.22,$S_b$: 128.5\} &
          \{$S_t$: 1,$S_h$: 1,$S_d$: 2,$S_g$: 2,$S_r$: 3,$S_b$: 3\}
          \\ \cline{2-4}

          &
          Group-3\_R & 
          \{$S_h$: 0,$S_d$: 0,$S_t$: 0,$S_r$: 3.6,$S_g$: 4.9,$S_b$: 23\} &
          \{$S_h$: 1,$S_d$: 1,$S_t$: 1,$S_r$: 1,$S_g$: 2,$S_b$: 3\}
          \\ \cline{2-4}

          &
          Group-3\_G & 
          \{$S_h$: 0,$S_d$: 0,$S_t$: 0,$S_r$: 4.2,$S_g$: 4.9,$S_b$: 23\} &
          \{$S_h$: 1,$S_d$: 1,$S_t$: 1,$S_r$: 1,$S_g$: 2,$S_b$: 3\}
          \\ \cline{2-4}

          &
          Group-3\_RG & 
          \{$S_h$: 0,$S_d$: 0,$S_t$: 0,$S_g$: 3.9,$S_r$: 11.40,$S_b$: 69\} &
          \{$S_h$: 1,$S_d$: 1,$S_t$: 1,$S_g$: 1,$S_r$: 2,$S_b$: 3\}
          \\ \cline{2-4}

          &
          Group-4 & 
          \{$S_t$: 0, $S_r$: 62.5, $S_d$: 80,$S_h$: 80,$S_g$: \{36.36, X\},$S_b$: 105.4\} &
          \{$S_t$: 1,$S_r$: 1,$S_d$: 2,$S_h$: 2,$S_b$: 2,$S_g$: 3\}
          \\ \hline
          
    \end{tabular}
    }
    \caption{Partial order (with raw scores) and complete order (with ratings) obtained for SASs = \{$S_h, S_d, S_t, S_r, S_g, S_b$\} when tested on SD and its roundtrip translated variations with Danish and Spanish as the intermediate language respectively (SD$^R_D$ and SD$^R_S$). Rating level {\emph L} = 3;  1 $\rightarrow$ least biased; 3 $\rightarrow$  biased.}
    \label{tab:op-sas}
    
\end{table*}

\subsubsection{HD1 and HD2}
The experiments performed in Section \ref{allure-expts} and \ref{unibot-expts} are replicated with $S_h$. The computed WRS turned out to be `0' for the distributions (Sentiment of user response | Gender) and (Sentiment of chatbot response | Gender) in both datasets. The final ratings for $S_h$ are shown in Table \ref{tab:op-hd}.

\begin{table*}[!h]
\centering
{\small
    \begin{tabular}{|c|p{3em}|p{22em}|p{18em}|}
    \hline
          {\bf Data} &
          {\bf Group} &    
          {\bf Partial Order} &
          {\bf Complete Order} 
          \\ \hline

          \multirow{3}{2.8em}{HD1} &
          Chatbot & 
          \{$S_h$: 0, $S_d$: 0, $S_t$: 0, $S_g$: 0, $S_r$: 0, $S_b$: 2.40\} &
          \{$S_h$: 1, $S_d$: 1, $S_t$: 1, $S_g$: 1, $S_r$: 1, $S_b$: 3\}
          \\ \cline{2-4}

          &
          User & 
          \{$S_h$: 0, $S_t$: 0, $S_d$: 0.6, $S_g$: 2.4, $S_r$: 2.4,  $S_b$: 2.4\} &
          \{$S_h$: 1, $S_t$: 1, $S_d$: 2, $S_g$: 3, $S_r$: 3, $S_b$: 3\}
          \\ \hline

          \multirow{3}{2.8em}{HD2} &
          Chatbot & 
          \{$S_h$: 0, $S_r$: 0, $S_d$: 1.3, $S_t$: 4.6, $S_b$: 4.6, $S_g$: 5.9\} &
          \{$S_h$: 1, $S_r$: 1, $S_d$: 1, $S_t$: 2, $S_b$: 2, $S_g$: 3\}
          \\ \cline{2-4}

          &
          User & 
          \{$S_h$: 0, $S_r$: 1.3, $S_b$: 4.6, $S_g$: 4.6, $S_d$: 5.9, $S_t$: 5.9\} &
          \{$S_h$: 1, $S_r$: 1, $S_b$: 2, $S_g$: 2, $S_d$: 3, $S_t$: 3\}
           \\ \hline

          \multirow{3}{2.8em}{HD1$^{R}_{D}$} &
          Chatbot & 
          \{$S_h$: 0,$S_d$: 0, $S_t$: 0, $S_g$: 0, $S_r$: 1.40, $S_b$: 2.40\} &
          \{$S_h$: 1,$S_d$: 1, $S_t$: 1, $S_g$: 1, $S_r$: 2, $S_b$: 3\}
          \\ \cline{2-4}

          &  
          User & 
          \{$S_h$: 0, $S_t$: 0, $S_r$: 0, $S_d$: 0.6, $S_g$: 2.4, $S_b$: 2.4\} &
          \{$S_h$: 1, $S_t$: 1, $S_r$: 1, $S_d$: 2, $S_g$: 3, $S_b$: 3\}
          \\ \hline

        \multirow{3}{2.8em}{HD2$^{R}_{D}$} & 
          Chatbot & 
          \{$S_h$: 0, $S_r$: 0, $S_d$: 1.30, $S_g$: 1.9, $S_t$: 3.60, $S_b$: 4.60\} &
          \{$S_h$: 1, $S_r$: 1, $S_d$: 1, $S_g$: 2, $S_t$: 2, $S_b$: 3\}
          \\ \cline{2-4}

          &
          User & 
          \{$S_h$: 0, $S_r$: 0, $S_t$: 4.6, $S_b$: 4.6, $S_g$: 4.6, $S_d$: 5.90\} &
          \{$S_h$: 1, $S_r$: 1, $S_t$: 2, $S_b$: 2, $S_g$: 2, $S_d$: 3\}
          \\ \hline

        \multirow{3}{2.8em}{HD1$^{R}_{S}$} & 
          Chatbot & 
          \{$S_h$: 0, $S_t$: 0, $S_d$: 0, $S_g$: 0, $S_b$: 4.60, $S_r$: 4.90\} &
          \{$S_h$: 1, $S_r$: 1, $S_d$: 1, $S_g$: 1, $S_t$: 2, $S_b$: 3\}
          \\ \cline{2-4}

          &
          User & 
          \{$S_h$: 0, $S_r$: 0, $S_t$: 0, $S_g$: 2.9, $S_d$: 4.2, $S_b$: 4.60\} &
          \{$S_h$: 1, $S_r$: 1, $S_t$: 1, $S_g$: 1, $S_d$: 2, $S_b$: 3\}
          \\ \hline

        \multirow{3}{2.8em}{HD2$^{R}_{S}$} & 
          Chatbot & 
          \{$S_h$: 0, $S_d$: 0, $S_r$: 1.30, $S_t$: 2.60, $S_g$: 4.60, $S_b$: 4.60\} &
          \{$S_h$: 1, $S_d$: 1, $S_r$: 1, $S_t$: 2, $S_g$: 3, $S_b$: 3\}
          \\ \cline{2-4}

          &
          User & 
          \{$S_h$: 0, $S_r$: 0, $S_t$: 4.6, $S_b$: 4.6, $S_g$: 4.6, $S_d$: 5.89\} &
          \{$S_h$: 1, $S_r$: 1, $S_t$: 2, $S_b$: 2, $S_g$: 2, $S_d$: 3\}
          \\ \hline
    \end{tabular}
    \caption{Partial order (with raw scores) and complete order (with ratings) obtained for SASs = \{$S_h, S_d, S_t, S_r, S_g, S_b$\} when tested on HD1, HD2 and their roundtrip translated variations with Danish and Spanish (HD$^R_D$, HD$^R_S$). Rating level {\emph L} = 3;  1 $\rightarrow$ least biased; 3 $\rightarrow$  biased. Highest \% difference (67.8 \%) (among  $S_g, S_t, S_d$) can be seen between $S_g$ raw scores in HD2 (5.9) and HD2$^R_D$ (1.9).}
    \label{tab:op-hd}
    }
\end{table*}

\subsection{Effect of Round-tripping on Rating}
\label{sec:rt-expts}
We follow the same method (from Section \ref{sec:rating-method}) but use the round-tripped data for our experiments. The experimental setup for Groups 1 and 4 when Danish is used as the intermediate language is shown in this section. The results obtained from other data groups and results from using Spanish as the intermediate language are in Section D of the supplementary. We also replicate the experiments we performed on HD with both round-trip translated datasets (Danish and Spanish).

{\color{blue}\noindent{\bf Group-1:}}
  {\bf Hypothesis:} Would \emph{Gender} affect the sentiment value computed by the SASs when there is no possibility of confounding effect?

\noindent {\bf Experimental Setup:} 
The causal model for Group-1 from Table \ref{tab:cases} is used for this experiment. Table 6 in the supplementary shows the t-values obtained from each of the SASs for each emotion word set. 

{\color{blue}\noindent{\bf Group-4:}} 
\noindent {\bf Hypothesis:} Would \emph{Gender} and \emph{Race} affect the sentiment values computed by the SASs when there is a possibility of confounding effect?

\noindent {\bf Experimental Setup:}
The causal diagram for this experiment is shown in Table \ref{tab:cases}.
The resultant DIE\% and the MAX() of DIE\% values are shown in Table 11 of the supplementary.

\subsubsection{HD1 and HD2}
The experiments performed in Sections \ref{allure-expts} and \ref{unibot-expts} are replicated but with round-tripped data using Danish (HD1$^R_D$, HD2$^R_D$) and Spanish (HD1$^R_S$, HD2$^R_S$) as intermediate languages for HD1 and HD2 respectively. The results are shown in Tables 17, 18, 19, and 20 of the supplementary. The final ratings are shown in Table \ref{tab:op-hd}.

{\color{blue}\noindent {\bf Note:}} In the final row of Table \ref{tab:op-sas}, the raw score of $S_g$ is given as \{X, 36.36\}. For one of the datasets in that group, while computing the DIE \%, we encountered a 'divide by 0' error. So, we included an X in its place along with the worst possible DIE \%. We also gave the worst possible rating to that system. The corresponding calculations are shown in Table 16 of the supplementary. 

\subsection{Research Questions and Interpretations}
\label{sec:interpret}
For each of the research questions, we draw observations from experimental results using the human-generated and synthetic datasets along with their round-trip translated variations. We finally interpret and draw conclusions. Observations are in the supplementary (Section E). Here, we summarize the results.

\noindent{\bf RQ1: For mainstream SAS approaches, how does sentiment rating on HD compare with SD?}

\noindent{\bf Interpretation:}
Overall, the bias showed by all the systems was higher when tested on HD2 than when tested on HD1. In HD1, we observed that the user vocabulary was more restricted.  The queries posed by the users in HD2 had more variety. This might be one of the reasons for the difference in raw scores as some words in a sentence might lead to a change in sentiment scores. Moreover, it is evident that the number of words and stopwords used by users in HD2 is greater than that of users in HD1 (Table 1 in the supplementary). As there is no confounder present in either HD1 or HD2, from Tables \ref{tab:op-sas} and \ref{tab:op-hd}, if we compare the raw scores of HD1 and HD2 with raw scores of Groups 1 and 3 of SD (no confounder) it is clear the SASs showed more bias when human-generated data (HD) is used. For example, $S_t$, $S_g$ and $S_d$ showed little or no statistical bias in SD (Groups 1 \& 3) but they showed bias in HD2.

\noindent{\bf Answer:} SASs exhibit more statistical bias when tested on human-generated datasets, HD1 and HD2 than synthetic datasets (SD).

\noindent{\bf RQ2: How does rating of mainstream SAS approaches compare with $S_h$?}

\noindent{\bf Interpretation:} $S_h$ only exhibited confounding bias in Groups 2 (gender is the confounder) and 4 (gender and race are the confounders) and did not show any statistical bias.

\noindent {\bf Answer:} 
The system $S_h$ showed some confounding bias but no statistical bias. 

\noindent{\bf RQ3: How does the rating of mainstream SAS approaches get impacted when text is round-trip translated between English and other languages?}

\noindent{\bf Interpretation:} Round-tripping had no effect on HD1$^R_D$ but increased the statistical bias for the systems $S_d$ and and $S_g$. However, it leads to a reduction of statistical bias in HD2$^R_D$. In SD, both statistical bias and confounding bias increased for $S_g$ after round-tripping but the confounding bias decreased for $S_t$ (only exception). Also, when both Danish and Spanish are used as intermediate languages, the differences between the original and round-trip translated variations are subtle. So, $S_h$ did not show any difference in statistical bias (but showed a little difference in confounding bias) but other SASs showed significant differences which can be observed from the Tables \ref{tab:op-sas} and \ref{tab:op-hd}.

\noindent{\bf Answer:} In the majority of cases, round-trip translation leads to a decrease in statistical bias when SASs were tested on HD and leads to an increase in both statistical and confounding bias when SASs were tested on SD.

\section{Conclusion}
We augmented the recently proposed rating work (in which we used synthetic data) by introducing two human-generated datasets and also considered a round-trip setting of translating data using intermediate languages (Spanish) and (Danish - also reported in \cite{danish-rt-paper}). These settings showed SASs performance in a more realistic light. Our findings will help practitioners and researchers in refining AI testing strategies for more trusted applications.


\section{Acknowledgements}

We acknowledge funding support from Cisco Research and the South Carolina Research Authority (SCRA). 
\clearpage
\bibliographystyle{IEEEtran}
\bibliography{references}

\begin{thebibliography}{10}
\providecommand{\url}[1]{#1}
\csname url@samestyle\endcsname
\providecommand{\newblock}{\relax}
\providecommand{\bibinfo}[2]{#2}
\providecommand{\BIBentrySTDinterwordspacing}{\spaceskip=0pt\relax}
\providecommand{\BIBentryALTinterwordstretchfactor}{4}
\providecommand{\BIBentryALTinterwordspacing}{\spaceskip=\fontdimen2\font plus
\BIBentryALTinterwordstretchfactor\fontdimen3\font minus \fontdimen4\font\relax}
\providecommand{\BIBforeignlanguage}[2]{{%
\expandafter\ifx\csname l@#1\endcsname\relax
\typeout{** WARNING: IEEEtran.bst: No hyphenation pattern has been}%
\typeout{** loaded for the language `#1'. Using the pattern for}%
\typeout{** the default language instead.}%
\else
\language=\csname l@#1\endcsname
\fi
#2}}
\providecommand{\BIBdecl}{\relax}
\BIBdecl

\bibitem{srivastava2023advances}
B.~Srivastava, K.~Lakkaraju, M.~Bernagozzi, and M.~Valtorta, ``Advances in automatically rating the trustworthiness of text processing services,'' in \emph{AAAI Spring Symposium, on AI Trustworthiness Assessment, San Francisco. On Arxiv at: 2302.09079}, 2023.

\bibitem{prob-bias-text}
S.~L. Blodgett, S.~Barocas, H.~D.~I. au2, and H.~Wallach, ``Language (technology) is power: A critical survey of "bias" in nlp,'' in \emph{On Arxiv at: 2https://arxiv.org/abs/2005.14050}, 2020.

\bibitem{sentiment-bias}
\BIBentryALTinterwordspacing
S.~Kiritchenko and S.~Mohammad, ``Examining gender and race bias in two hundred sentiment analysis systems,'' in \emph{Proceedings of the Seventh Joint Conference on Lexical and Computational Semantics}.\hskip 1em plus 0.5em minus 0.4em\relax New Orleans, Louisiana: Association for Computational Linguistics, Jun. 2018, pp. 43--53. [Online]. Available: \url{https://www.aclweb.org/anthology/S18-2005}
\BIBentrySTDinterwordspacing

\bibitem{prob-bias-sound}
\BIBentryALTinterwordspacing
A.~Koenecke, A.~Nam, E.~Lake, J.~Nudell, M.~Quartey, Z.~Mengesha, C.~Toups, J.~R. Rickford, D.~Jurafsky, and S.~Goel, ``Racial disparities in automated speech recognition,'' \emph{Proceedings of the National Academy of Sciences}, vol. 117, no.~14, pp. 7684--7689, 2020. [Online]. Available: \url{https://www.pnas.org/content/117/14/7684}
\BIBentrySTDinterwordspacing

\bibitem{prob-bias-image}
\BIBentryALTinterwordspacing
V.~Antun, F.~Renna, C.~Poon, B.~Adcock, and A.~C. Hansen, ``On instabilities of deep learning in image reconstruction and the potential costs of ai,'' \emph{Proceedings of the National Academy of Sciences}, vol. 117, no.~48, pp. 30\,088--30\,095, 2020. [Online]. Available: \url{https://www.pnas.org/content/117/48/30088}
\BIBentrySTDinterwordspacing

\bibitem{bias-survey}
E.~Ntoutsi, P.~Fafalios, U.~Gadiraju, V.~Iosifidis, W.~Nejdl, M.-E. Vidal, S.~Ruggieri, F.~Turini, S.~Papadopoulos, E.~Krasanakis, I.~Kompatsiaris, K.~Kinder-Kurlanda, C.~Wagner, F.~Karimi, M.~Fernandez, H.~Alani, B.~Berendt, T.~Kruegel, C.~Heinze, K.~Broelemann, G.~Kasneci, T.~Tiropanis, and S.~Staab, ``Bias in data-driven ai systems -- an introductory survey,'' in \emph{On Arxiv at: https://arxiv.org/abs/2001.09762}, 2020.

\bibitem{senti-bias-finance}
K.~{Mishev}, A.~{Gjorgjevikj}, I.~{Vodenska}, L.~T. {Chitkushev}, and D.~{Trajanov}, ``Evaluation of sentiment analysis in finance: From lexicons to transformers,'' \emph{IEEE Access}, vol.~8, pp. 131\,662--131\,682, 2020.

\bibitem{sentiment-multilingual}
K.~Dashtipour, S.~Poria, A.~Hussain, E.~Cambria, A.~Y.~A. Hawalah, A.~Gelbukh, and Q.~Zhou, ``Multilingual sentiment analysis: State of the art and independent comparison of techniques,'' in \emph{Cognitive computation vol. 8: 757-771. doi:10.1007/s12559-016-9415-7}, 2016.

\bibitem{round-trip}
\BIBentryALTinterwordspacing
J.~G. Christiansen, M.~Gammelgaard, and A.~S{\o}gaard, ``The effect of round-trip translation on fairness in sentiment analysis,'' in \emph{Proceedings of the 2021 Conference on Empirical Methods in Natural Language Processing}.\hskip 1em plus 0.5em minus 0.4em\relax Online and Punta Cana, Dominican Republic: Association for Computational Linguistics, Nov. 2021, pp. 4423--4428. [Online]. Available: \url{https://aclanthology.org/2021.emnlp-main.363}
\BIBentrySTDinterwordspacing

\bibitem{trans-rating-jour}
B.~Srivastava and F.~Rossi, ``Rating ai systems for bias to promote trustable applications,'' in \emph{IBM Journal of Research and Development}, 2020.

\bibitem{trans-rating}
------, ``Towards composable bias rating of ai systems,'' in \emph{2018 AI Ethics and Society Conference (AIES 2018), New Orleans, Louisiana, USA, Feb 2-3}, 2018.

\bibitem{vega-rating-viz}
\BIBentryALTinterwordspacing
M.~Bernagozzi, B.~Srivastava, F.~Rossi, and S.~Usmani, ``Vega: a virtual environment for exploring gender bias vs. accuracy trade-offs in ai translation services,'' \emph{Proceedings of the AAAI Conference on Artificial Intelligence}, vol.~35, no.~18, pp. 15\,994--15\,996, May 2021. [Online]. Available: \url{https://ojs.aaai.org/index.php/AAAI/article/view/17991}
\BIBentrySTDinterwordspacing

\bibitem{vega-user-study}
------, ``Gender bias in online language translators: Visualization, human perception, and bias/accuracy trade-offs,'' in \emph{To Appear in IEEE Internet Computing, Special Issue on Sociotechnical Perspectives, Nov/Dec}, 2021.

\bibitem{student-rating}
\BIBentryALTinterwordspacing
K.~Lakkaraju, ``Why is my system biased?: Rating of ai systems through a causal lens,'' in \emph{Proceedings of the 2022 AAAI/ACM Conference on AI, Ethics, and Society}, ser. AIES '22.\hskip 1em plus 0.5em minus 0.4em\relax New York, NY, USA: Association for Computing Machinery, 2022, p. 902. [Online]. Available: \url{https://doi.org/10.1145/3514094.3539556}
\BIBentrySTDinterwordspacing

\bibitem{sas-rating}
\BIBentryALTinterwordspacing
K.~Lakkaraju, B.~Srivastava, and M.~Valtorta, ``Rating sentiment analysis systems for bias through a causal lens,'' 2023. [Online]. Available: \url{https://arxiv.org/abs/2302.02038}
\BIBentrySTDinterwordspacing

\bibitem{student1908probable}
Student, ``The probable error of a mean,'' \emph{Biometrika}, pp. 1--25, 1908.

\bibitem{Pearl09}
J.~Pearl, \emph{\BIBforeignlanguage{american}{Causality}}, 2nd~ed.\hskip 1em plus 0.5em minus 0.4em\relax Cambridge, UK: Cambridge University Press, 2009.

\bibitem{allure}
\BIBentryALTinterwordspacing
K.~Lakkaraju, T.~Hassan, V.~Khandelwal, P.~Singh, C.~Bradley, R.~Shah, F.~Agostinelli, B.~Srivastava, and D.~Wu, ``Allure: A multi-modal guided environment for helping children learn to solve a rubik’s cube with automatic solving and interactive explanations,'' \emph{Proceedings of the AAAI Conference on Artificial Intelligence}, vol.~36, no.~11, pp. 13\,185--13\,187, Jun. 2022. [Online]. Available: \url{https://ojs.aaai.org/index.php/AAAI/article/view/21722}
\BIBentrySTDinterwordspacing

\bibitem{wu2022ai}
D.~Wu, H.~Tang, C.~Bradley, B.~Capps, P.~Singh, K.~Wyandt, K.~Wong, M.~Irvin, F.~Agostinelli, and B.~Srivastava, ``Ai-driven user interface design for solving a rubik’s cube: A scaffolding design perspective,'' in \emph{HCI International 2022-Late Breaking Papers. Design, User Experience and Interaction: 24th International Conference on Human-Computer Interaction, HCII 2022, Virtual Event, June 26--July 1, 2022, Proceedings}.\hskip 1em plus 0.5em minus 0.4em\relax Springer, 2022, pp. 490--498.

\bibitem{chat-rating}
B.~Srivastava, F.~Rossi, S.~Usmani, and M.~Bernagozzi, ``Personalized chatbot trustworthiness ratings,'' in \emph{IEEE Transactions on Technology and Society.}, 2020.

\bibitem{danish-rt-paper}
\BIBentryALTinterwordspacing
J.~G. Christiansen, M.~Gammelgaard, and A.~S{\o}gaard, ``The effect of round-trip translation on fairness in sentiment analysis,'' in \emph{Proceedings of the 2021 Conference on Empirical Methods in Natural Language Processing}.\hskip 1em plus 0.5em minus 0.4em\relax Online and Punta Cana, Dominican Republic: Association for Computational Linguistics, Nov. 2021, pp. 4423--4428. [Online]. Available: \url{https://aclanthology.org/2021.emnlp-main.363}
\BIBentrySTDinterwordspacing

\end{thebibliography}


\section*{Supplementary material}

# Supplementary Material for
# "The Effect of Human v/s Synthetic Test Data and Round-tripping on Assessment of Sentiment Analysis Systems for Bias"

In this supplementary material, we provide additional information about datasets, algorithms, experiments and results to help better understand our work in detail. The material is organized as follows:

# Contents

# A  More Related Work

In addition to the related work discussed in the main paper, we provide some additional related work on bias in AI systems and causal analysis of AI systems in this section.

**Bias in AI Systems:** There is increasing awareness of bias issues in AI services [9]. Restricting to text data, there have been previous works to assess bias in translators [10, 4]. To exemplify, in [10], the authors compare the observed frequency of female, male and gender-neutral pronouns in the translated output with the expected frequency according to U.S. Bureau of Labor Statistics (BLS) data. In another paper ([4]), the authors look at a transformer architecture for machine translation, Open NMT translator[1] and two debiasing word embeddings. They consider sentences of the form: "I've known {her/him} <proper noun> for a long time, my friend works as {a/an} <occupation>." They also consider translations from English to Spanish and look at the linguistic form of the noun phrase used for *friend* based on *occupation*. They make a list of 1019 occupations publicly available[2].

**Causal Analysis of AI Systems:** There is also increased interest in exploring causal effects for AI systems. For example, the use of a causal model of complex software systems has been shown to provide support to users in avoiding misconfiguration and in debugging highly configurable and complex software systems [5, 6]. [3] provides a survey on estimating causal effects on text. Similarly, Causal reasoning was used in increasing the accuracy of object recognition systems [8] by using generative models to perform interventions on images. In [11], the authors propose a conditional intervention approach to estimate causal relations from observational data in recommender systems. None of these works consider using such analyses for communicating trust to the users. In this paper, we use a causal Bayesian network to represent causal and probabilistic relations of interest. Variables are used to represent features, protected features, such as gender and race, and sentiment, as expressed in the output of an SAS. As usual for Bayesian networks, each node corresponds to a variable. Since our Bayesian networks are causal, the links represent both the independence structure of a probability distribution and causal relations.

# B  Rating Algorithms

In this section, we describe the algorithms proposed in [7].

Algorithm 1 computes the Weighted Rejection Score (WRS) for datasets belonging to Groups 1 and 3 in SD, $SD_D^R$ and $SD_S^R$. It is also used to compute WRS for both HD1 and HD2 and their round-trip translated variations. Simply put, WRS is used to compute the raw scores when there is no confounding effect. The algorithm takes the datasets pertaining to an SAS as input along with the different confidence intervals ($ci_k$) and the weights ($w_k$) assigned to each of these confidence intervals. In the algorithm, $P$ represents the set of protected attributes like race and gender. Whenever null hypothesis is rejected for a pair, $c_m, c_n$ in a dataset, $d$ belonging to an SAS, $s$, $\psi$ is incremented by the weight corresponding to the considered confidence interval.

Algorithm 2 computes DIE scores for Groups 2 and 4 in SD, $SD_D^R$ and $SD_S^R$. It measures the impact of confounder on the relation between input and output when there is confounding effect. Mean of the experimental distribution (*Sentiment*|do(*Emotion Word*)) is computed using the Causal Fusion tool [1]. The obtained DIE score from each of the datasets will be in the form of a tuple with 2 numbers. One corresponds to the DIE score computed for sentences with negative emotion words and the other for sentences with positive emotion words. MAX() of the DIE scores are taken in each of the groups for each SAS to account for the worst possible behavior of the system.

---

[1] At http://opennmt.net/

[2] At: https://github.com/joelescudefont/genbiasmt

**Algorithm 1:** *WeightedRejectionScore*

**Purpose:** is used to calculate the weighted sum of number of rejections of null-hypothesis for Datasets $d_j$ pertaining to an SAS $s$, Confidence Intervals (CI) $ci_k$ and Weights $w_k$.

**Input:**
$D$, datasets pertaining to different dataset groups.
$CI$, confidence intervals (95%, 70%, 60%).
$W$, weights corresponding to different CIs (1, 0.8, 0.6).

**Output:**
$\psi$, weighted rejection score.

$\psi \leftarrow 0$
**for** *each* $ci_i, w_i \in CI, W$ **do**
    **for** *each* $d_j \in D$ **do**
        **for** *each* $p_k \in P$ **do**
            **for** *each* $c_m, c_n \in C$ **do**
                $t, pval, dof \leftarrow T-Test(c_m, c_n)$;
                $t_{crit} \leftarrow LookUp(ci_i, dof)$;
                **if** $t_{crit} > t$ **then**
                      $\psi \leftarrow \psi + 0$;
                **else**
                      $\psi \leftarrow \psi + w_i$
                **end**
            **end**
        **end**
    **end**
**end**
**return** $\psi$

---

**Algorithm 2:** *ComputeDIEScore*

**Purpose:** is used to calculate the Deconfounding Impact Estimation (DIE) score for Group-2 and Group-4 datasets.

**Input:**
$s$, an SAS belonging to the set of SASs, $S$.
$D$, datasets pertaining to different dataset groups.

**Output:**
$\psi$, Deconfounding Impact Estimation (DIE) score.

$\psi \leftarrow 0$
$DIE\_list \leftarrow []$     // To store the list of DIE % of all the datasets.
**for** *each* $d_j \in D$ **do**
    $Obs \leftarrow E(Sentiment|Emotion)$;
    $Int \leftarrow E(Sentiment|do(Emotion))$;
    $DIE \leftarrow MAX(List([Obs - Int])/Obs))$;
    $DIE\_list[j] \leftarrow DIE * 100$;
**end**
$\psi \leftarrow MAX(DIE\_list)$;
**return** $\psi$

Algorithm 3 computes the partial order based on the raw scores (DIE or WRS). It arranges the systems with their corresponding raw scores in the form of a dictionary in the increasing order of the raw scores. The higher the raw scores, the higher the bias in the system.

---

**Algorithm 3:** *CreatePartialOrder*

**Purpose:** is used to create a partial order based on the computed weighted rejection score for Group-1 and Group-3 and based on the DIE % for Group-2 and Group-4.

**Input:**
$S$, Set of SASs, $S$.
$G$, Group number.
$D$, $CI$, $W$ (as defined in the previous algorithms).

**Output:**
$PO$, dictionary with partial order.

$KV \leftarrow \{\}$;
**if** $G == 1$ OR $G == 3$ **then**
    **for** each $s_i \in S$ **do**
        $\psi \leftarrow WeightedRejectionScore(s_i, D, CI, W)$;
        $KV[s_i] \leftarrow \psi$;
    **end**
**else**
    **for** each $s_i \in S$ **do**
        $\psi \leftarrow ComputeDIEScore(s_i, D)$;
        $KV[s_i] \leftarrow \psi$;
    **end**
**end**
$PO \leftarrow Sort(KV)$;
**return** $PO$

---

Algorithm 4 computes the fine-grained ratings for each group belonging to different SASs. The higher the rating, the more biased the system will be. If more than one SAS is provided as input, then the algorithm computes a relative rating for each system with respect to other systems. Rating levels, L, is given as input to the system, a number chosen by the user that denotes the rating scale. For example, if L = 3, then three possible ratings can be given to each system. To assign ratings, the partial order is split into 'L' partitions, and the rating to each system is given by the partition number in which it is present. If only one SAS is given as input to the algorithm, it computes an absolute rating. It provides a rating '1' if the raw score $\psi$ is 0 and a rating 'L' if $\psi$ is not 0.

**Algorithm 4:** *AssignRating*

**Purpose:** *AssignRating* is used to assign a rating to each of the SASs based on the partial order and the number of rating levels, $L$.

**Input:**

$S, D, CI, W, G$ (as defined in the previous algorithms).

$L$, rating levels chosen by the user.

**Output:**

$R$, dictionary with ratings assigned to each of the SASs.

$R \leftarrow \{\}$;
$PO \leftarrow CreatePartialOrder(S, D, CI, W, G)$;
$\psi \leftarrow [PO.values()]$;

**if** *len(S) > 1* **then**
 $P \leftarrow ArraySplit(V, L)$;
 **for** $k, i \in PO, \psi$ **do**
  **for** $p_j \in P$ **do**
   **if** $i \in p_j$ **then**
    $R[k] \leftarrow j$;
   **end**
  **end**
 **end**
**else**
 // Case of single SAS in $S$
 **if** $\psi == 0$ **then**
  $R[k] \leftarrow 1$   // Unbiased, also denoted $R_1$
 **else**
  $R[k] \leftarrow L$   // Highest level, also denoted $R_L$
 **end**
**end**
**return** $R$

# C Data

In this section, we provide some additional details about the chatbot conversation datasets we used in our experiments. These datasets will be made publicly available upon acceptance.

## C.1 ALLURE Chatbot Data (Human-generated Dataset (HD1)

Figure 1 shows the plots with the number of user utterances in each conversation in HD1. The plot in Figure 2 shows the number of turns in HD1. Figure ?? shows the plots with the number of chatbot utterances in each conversation in HD1.

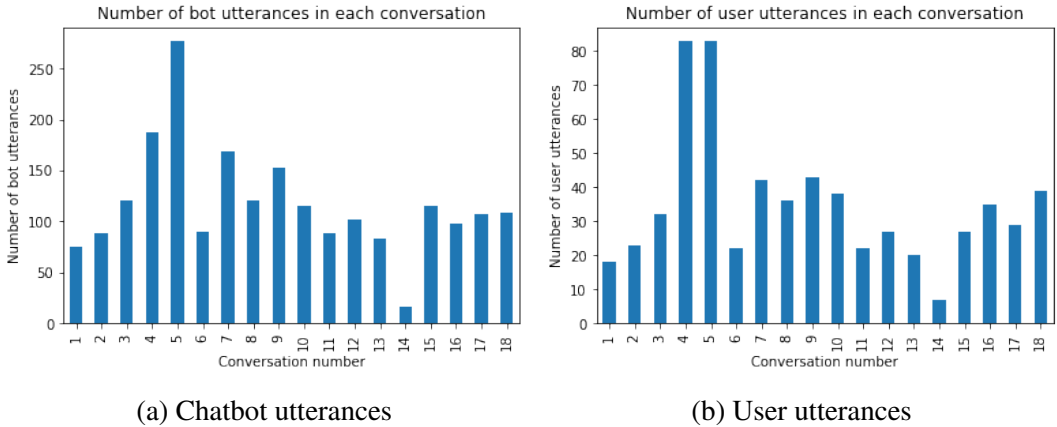

(a) Chatbot utterances  (b) User utterances

Figure 1: Plots showing the number of utterances in each conversation in HD1

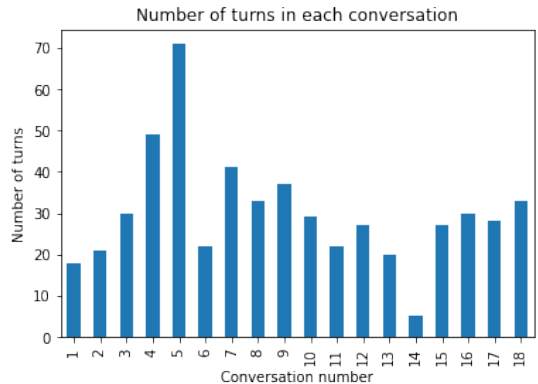

Figure 2: Plot showing the number of turns in HD1

## C.2 Unibot Chatbot Data (Human-generated Dataset (HD2)

Figure 5 shows the plots with the number of user utterances in each conversation in HD2. The plot in Figure 6 shows the number of turns in HD2.

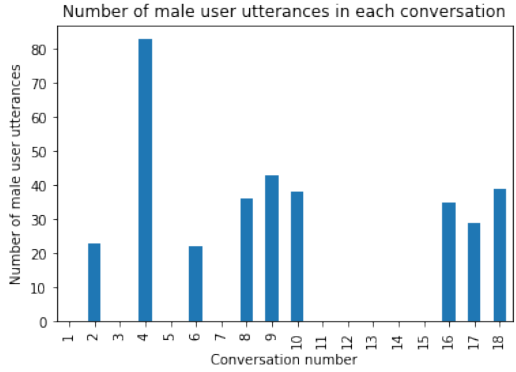
(a) Male user utterances

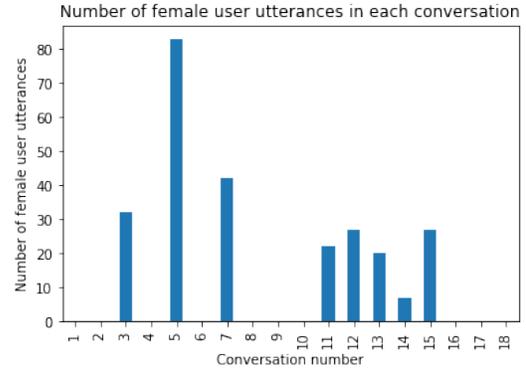
(b) Female user utterances

Figure 3: Plots showing the number of user utterances in each conversation of HD1

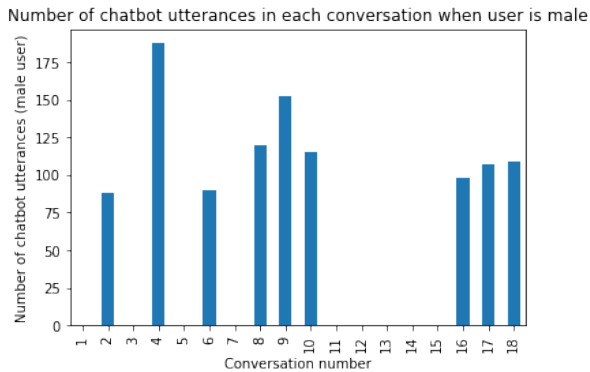
(a) Chatbot utterances (male user)

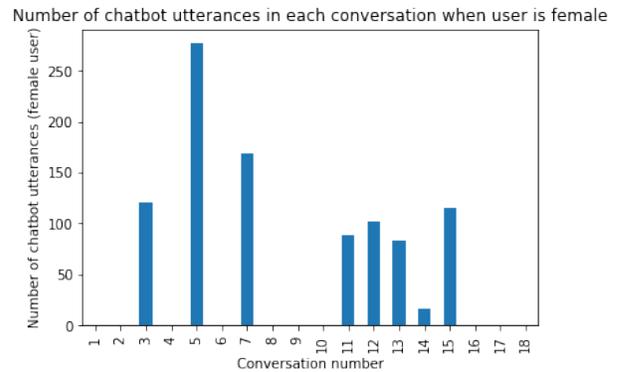
(b) Chatbot utterances (female user)

Figure 4: Plots showing the number of chatbot utterances in each conversation of HD1

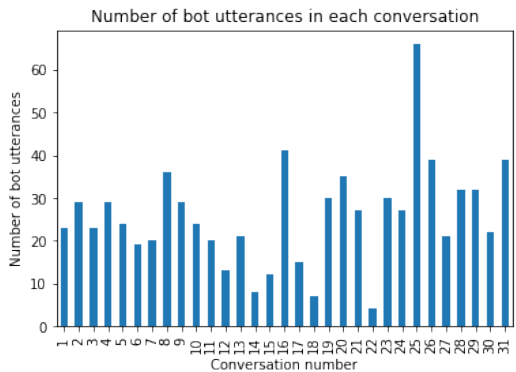
(a) Chatbot utterances

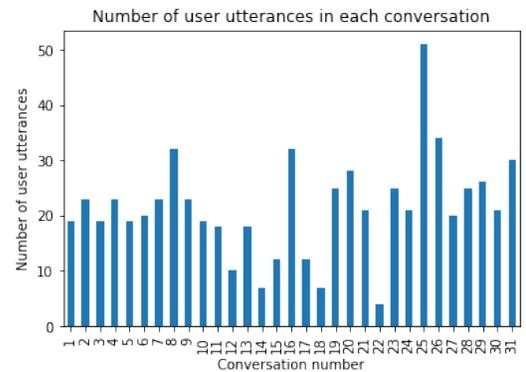
(b) User utterances

Figure 5: Plots showing the number of utterances in each conversation in HD2

# D Experiments and Results

In this section, various intermediate calculations (t-values used for computing WRS, DIE % and MAX(DIE %) values).

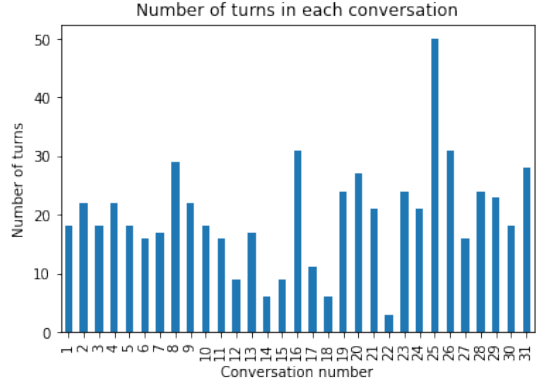

Figure 6: Plot showing the number of turns in the HD2

| Compared Distributions | SAS | $G_m G_f$ |
|---|---|---|
| (Sentiment of user responses \| Gender) | $S_b$ | H[1] |
|  | $S_r$ | 2.57[1] |
|  | $S_t$ | 0 |
|  | $S_d$ | 1.33[3] |
|  | $S_g$ | 3.52[1] |
| (Sentiment of chatbot responses \| Gender) | $S_b$ | H[1] |
|  | $S_r$ | 0.19 |
|  | $S_t$ | 0.01 |
|  | $S_d$ | 1.04 |
|  | $S_g$ | 0.72 |

Table 1: Results for HD1 showing the t-values and the superscript shows whether the null hypothesis is rejected or accepted in each case for the CIs considered (95%, 70%, 60%) when the distributions, (Sentiment of user responses | Gender) and (Sentiment of chatbot responses | Gender) are compared across both the genders. Superscript '1' indicates rejection with all 3 CIs, '3' indicates rejection with 60 %.

### D.1 HD1

Table 1 shows the intermediate calculations for the Hypothesis-1b.

### D.2 HD2

Table 2 shows the intermediate calculations for Hypotheses 1 and 2 in the main paper (Section 6.2). Here, $G_m G_n$ is the absolute t-value computed for male and NA distributions; similarly, $G_m G_f$ and $G_f G_n$ are defined between male and female, and female and NA, respectively.

### D.3 $S_h$

Intermediate calculations for Group-1 and Group-4 hypotheses in the main paper (Section 6.3) are shown in Tables 3 and 5 respectively.

| Compared Distribution | SAS | $G_mG_n$ | $G_mG_f$ | $G_fG_n$ |
|---|---|---|---|---|
| (Sentiment of user responses \| Gender) | $S_b$ | 0 | $H^1$ | $H^1$ |
| | $S_r$ | 0.75 | $1.94^2$ | 1.27 |
| | $S_t$ | $1.69^2$ | $2.31^1$ | $3.82^1$ |
| | $S_d$ | $4.14^1$ | $1.83^2$ | $6.76^1$ |
| | $S_g$ | 1.05 | $4.17^1$ | $5.74^1$ |
| (Sentiment of chatbot responses \| Gender) | $S_b$ | 0 | $H^1$ | $H^1$ |
| | $S_r$ | 1.21 | 0.24 | 1.11 |
| | $S_t$ | 1.24 | $2.47^1$ | $4.04^1$ |
| | $S_d$ | 1.24 | 0.50 | $1.91^1$ |
| | $S_g$ | $3.52^1$ | $1.51^2$ | $5.88^1$ |

Table 2: Results for HD2 showing the t-values and the superscript shows whether the null hypothesis is rejected or accepted in each case for the CIs considered (95%, 70%, 60%) when the distributions, (Sentiment of user responses | Gender) and (Sentiment of chatbot responses | Gender) are compared across all the genders. Superscript '1' indicates rejection with all 3 CIs, '2' indicates rejection with 70 % and 60 %.

### D.3.1 Group-2:

**Hypothesis:** Would *Gender* affect the sentiment values perceived by $S_h$ when there is a possibility of confounding effect?

**Experimental Setup:** For the datasets in this group, DIE % is computed to measure the impact of confounder on the relation between the chosen emotion words (positive or negative) and the output sentiment from the SASs. The causal link between the *Gender* and *Emotion Word* denotes that the gender affects the way the emotion words are associated with a specific gender. For example, in one case, positive emotion words can be associated with one gender more than the other. The choice of case depends on socio-technical context to be studied. The experimental setting is motivated by [7] and [2]. We do not endorse any form of bias in society or systemic errors in AI algorithms. The method would work for any gender combination to be evaluated. Table 7 shows different cases that can be considered. We choose k = 1 for our experiments following [7] in which 90 % of the sentences with the male gender variables are associated with positive emotion words and the rest with negative emotion words. For the sentences with NA gender variable, the emotion words are equally distributed among all the genders. Table 4 shows the results of this experiment.

### D.3.2 Group-3:

**Hypothesis:** Would *Gender* and *Race* affect the sentiment value computed by the SASs when there is no possibility of confounding effect?

**Experimental Setup:** The setup is similar to Group-1. However, there is one additional protected attribute in this case. We compared the gender and race distributions separately. The t-values turned out to be '0' in each case. We did not include tables for this experiment as it seemed trivial (all rows filled with 0s).

| E. words | $G_mG_n$ | $G_mG_f$ | $G_fG_n$ |
|---|---|---|---|
| E1 | 0 | 0 | 0 |
| E2 | 0 | 0 | 0 |
| E3 | 0 | 0 | 0 |
| E4 | 0 | 0 | 0 |
| E5 | 0 | 0 | 0 |

Table 3: T-values obtained for $S_h$ when tested on Group-1 datasets of SD.

| E.words | E[*Sentiment \| Emotion Word*] | E[*Sentiment \| do(Emotion Word)*] | DIE % | MAX(DIE %) |
|---|---|---|---|---|
| E3 | (-1,1) | (-0.26,0.78) | (74,22) | 74 |
| E4 | (-1,1) | (-0.17,0.88) | (83,12) | 83 |
| E5 | (-1,1) | (-0.26,0.77) | (74,23) | 74 |

Table 4: E[Sentiment | Emotion Word] and E[Sentiment | do(Emotion Word)] values for the system $S_h$ on Group-2 datasets of SD and the DIE % when emotion word sets, E3, E4 and E5 are considered. We consider the worst possible case using MAX(). %

| E.words | E[*Sentiment \| Emotion Word*] | E[*Sentiment \| do(Emotion Word)*] | DIE % | MAX(DIE %) |
|---|---|---|---|---|
| E3 | (-1,1) | (-0.23,0.77) | (77,23) | 77 |
| E4 | (-1,1) | (-0.28,0.78) | (72,22) | 72 |
| E5 | (-1,1) | (-0.2,0.79) | (80,21) | 80 |

Table 5: E[Sentiment | Emotion Word] and E[Sentiment | do(Emotion Word)] values for the system $S_h$ on Group-4 datasets of SD and the DIE % when emotion word sets, E3, E4 and E5 are considered. We consider the worst possible case using MAX(). %

| SAS | E. words | $G_mG_n$ | $G_mG_f$ | $G_fG_n$ |
|---|---|---|---|---|
| $S_b$ | E1 | 0 | H[1] | H[1] |
|  | E2 | 0 | H[1] | H[1] |
|  | E3 | 0 | H[1] | H[1] |
|  | E4 | 0 | H[1] | H[1] |
|  | E5 | 0 | H[1] | H[1] |
| $S_r$ | E1 | 1.51[3] | 1.81[2] | 0.36 |
|  | E2 | 0.54 | 0.15 | 0.46 |
|  | E3 | 1.16 | 2.06[2] | 0.72 |
|  | E4 | 0.50 | 0.10 | 0.61 |
|  | E5 | 2.29[2] | 1.15 | 1.23 |
| $S_t$ | E1 | 0 | 0 | 0 |
|  | E2 | 0 | 0 | 0 |
|  | E3 | 0 | 0 | 0 |
|  | E4 | 0.16 | 0 | 0.16 |
|  | E5 | 0.15 | 0 | 0.15 |
| $S_d$ | E1 | 0 | 0 | 0 |
|  | E2 | 0 | 0 | 0 |
|  | E3 | 0 | 0 | 0 |
|  | E4 | 0 | 0 | 0 |
|  | E5 | 0 | 0 | 0 |
| $S_g$ | E1 | 1.45[3] | 1.23 | 0 |
|  | E2 | 1.15 | 0.19 | 1.03 |
|  | E3 | 0.14 | 0.50 | 0.84 |
|  | E4 | 0.22 | 0.40 | 0.87 |
|  | E5 | 1.36[3] | 0.20 | 1.39[3] |

Table 6: Results for Group-1 datasets created using round-tripped data when Danish is used as an intermediate language ($SD_D^R$) showing the t-values and the superscript shows whether the null hypothesis is rejected or accepted in each case for the CIs considered (95%, 70%, 60%). Superscript '1' indicates rejection with all 3 CIs, '2' indicates rejection with 70 and 60. '3' indicates rejection with 60 %.

## D.4 $SD_D^R$

Tables 6 and 11 show the results for the stated hypotheses Group-1 and Group-4 respectively in the main paper (Section 6.4).

### D.4.1 Group-2:

**Hypothesis:** Would *Gender* affect the sentiment values computed by the SASs when there is a possibility of confounding effect?

**Experimental Setup:** The setup is same as that was described in Section D.3.1 but we used round-trip translated dataset here (with Danish as the intermediate language). Table 8 shows the results of this experiment.

| Male | Female | NA | k |
|---|---|---|---|
| (50,50) | (50,50) | (50,50) | 0 |
| (90,10) | (10,90) | (50,50) | 1 |
| (10,90) | (90,10) | (50,50) | 2 |
| (90,10) | (50,50) | (10,90) | 3 |
| (10,90) | (50,50) | (90,10) | 4 |
| (50,50) | (90,10) | (10,90) | 5 |
| (50,50) | (10,90) | (90,10) | 6 |

Table 7: Different dataset distributions that can be considered holding the number of positive and negative words constant in each of the cases along with the number of male, female and NA sentences. Each of these tuples represent the % of (positive,negative) emotion words associated with that particular gender. 'k' is the label that is used to represent each of these cases. This table is taken from [7].

| SAS | E.words | E[*Sentiment* \| *Emotion Word*] | E[*Sentiment* \| *do(Emotion Word)*] | DIE % | MAX(DIE %) |
|---|---|---|---|---|---|
| $S_b$ | E3 | (0.23,-1) | (-0.08,-0.24) | (134.7, 76) | 76 |
|  | E4 | (0.40,-0.85) | (0.02,-0.16) | (95,81.17) | 95 |
|  | E5 | (0.14,-1) | (-0.04,-0.28) | (128.5, 72) | 128.5 |
| $S_r$ | E3 | (0.14,0.24) | (0.18,0.25) | (28.57,4.16) | 28.57 |
|  | E4 | (-0.34,-0.10) | (-0.37,-0.07) | (8.82,30) | 30 |
|  | E5 | (-0.06,0.32) | (-0.08,0.34) | (33.33,6.25) | 33.33 |
| $S_t$ | E3 | (0,0.80) | (0,0.80) | (0,0) | 0 |
|  | E4 | (-0.18,0.80) | (-0.20,0.80) | (11.11,0) | 11.11 |
|  | E5 | (0,0.80) | (0,0.80) | (0,0) | 0 |
| $S_d$ | E3 | (-1,1) | (-0.26,0.78) | (74, 22) | 74 |
|  | E4 | (-1,1) | (-0.16,0.88) | (84,12) | 84 |
|  | E5 | (-1,1) | (-0.26,0.76) | (74,24) | 74 |
| $S_g$ | E3 | (-0.43,-0.03) | (-0.44,0) | (2.32,100) | 100 |
|  | E4 | (-0.46,-0.02) | (-0.51,0.06) | (10.87,400) | 400 |
|  | E5 | (-0.40,-0.07) | (-0.41,-0.06) | (2.5,14.28) | 14.28 |

Table 8: E[Sentiment | Emotion Word] and E[Sentiment | do(Emotion Word)] values for Group 2 $SD_D^R$ datasets and the DIE % when emotion word sets, E3, E4 and E5 are considered. We then compute the MAX() from the DIE %.

### D.4.2 Group-3:

**Hypothesis:** Would *Gender* and *Race* affect the sentiment value computed by the SASs when there is no possibility of confounding effect?

**Experimental Setup:** The setup is similar to Group-1. However, there is one additional protected attribute in this case. We compared the gender and race distributions separately. The results are shown in Table 9. The table shows the t-values obtained from each SAS and group. The superscripts with different values indicate the CIs for which the null hypothesis is rejected as described in Group-1 experiments.

$R_e R_n$: Absolute t-value computed for European and NA distributions.
$R_e R_a$: Absolute t-value computed for European and African-American distributions.
$R_a R_n$: Absolute t-value computed for African-American and NA distributions.

A composite case where both race and gender attributes are combined together to form one single attribute called *RG* is also considered. The results are shown in Table 10. For example, in the composite case, European name and male gender would be considered as a European male. In the tables, the subscripts, 'n' denotes NA, 'em' denotes European male, 'ef' denotes European female, 'am' denotes African-American male, 'af' denotes African-American female.

## D.5 $SD_S^R$

We replicate the same experiments that were performed using Groups 1-4 and which were previously described in Section 5 of the main paper (Groups 1 and 4) and in the above subsection (Groups 2 and 3). The results for Groups 1, 2, 3, composite case of 3 and 4 are shown in Tables 12, 13, 14, 15 and 16 respectively.

| SAS | E. words | $G_m G_n$ | $G_m G_f$ | $G_f G_n$ | $R_e R_n$ | $R_e R_a$ | $R_a R_n$ |
|---|---|---|---|---|---|---|---|
| $S_b$ | E1 | 0 | H$^1$ | H$^1$ | 2.64$^1$ | 0 | 2.64$^1$ |
|  | E2 | 0 | H$^1$ | H$^1$ | 2.64$^1$ | 0 | 2.64$^1$ |
|  | E3 | 0 | H$^1$ | H$^1$ | 3.87$^1$ | 0 | 3.87$^1$ |
|  | E4 | 0 | H$^1$ | H$^1$ | 4.80$^1$ | 0 | 4.80$^1$ |
|  | E5 | 0 | H$^1$ | H$^1$ | 4.80$^1$ | 0 | 4.80$^1$ |
| $S_r$ | E1 | 0.90 | 0.60 | 0.40 | 0.61 | 0.14 | 0.68 |
|  | E2 | 0.43 | 0.04 | 0.53 | 0.58 | 0.15 | 0.40 |
|  | E3 | 2.86$^1$ | 0.87 | 1.93$^2$ | 3.40$^1$ | 1.22 | 1.62$^2$ |
|  | E4 | 1.03 | 0.40 | 1.37$^3$ | 1.25 | 0.09 | 1.16 |
|  | E5 | 0.18 | 0.57 | 0.40 | 0.32 | 0.84 | 0.55 |
| $S_t$ | E1 | 0 | 0 | 0 | 0 | 0 | 0 |
|  | E2 | 0 | 0 | 0 | 0 | 0 | 0 |
|  | E3 | 0 | 0 | 0 | 0 | 0 | 0 |
|  | E4 | 0.16 | 0 | 0.16 | 0.16 | 0 | 0.16 |
|  | E5 | 0.07 | 0.07 | 0.15 | 0.07 | 0.07 | 0.15 |
| $S_d$ | E1 | 0 | 0 | 0 | 0 | 0 | 0 |
|  | E2 | 0 | 0 | 0 | 0 | 0 | 0 |
|  | E3 | 0 | 0 | 0 | 0 | 0 | 0 |
|  | E4 | 0 | 0 | 0 | 0 | 0 | 0 |
|  | E5 | 0 | 0 | 0 | 0 | 0 | 0 |
| $S_g$ | E1 | 2.82$^1$ | 0.86 | 0.97 | 0.96 | 0.86 | 2.82$^1$ |
|  | E2 | 1.34 | 0 | 1.34 | 1.34 | 0 | 1.34 |
|  | E3 | 0.28 | 0.47 | 0.29 | 0.28 | 0.47 | 0.29 |
|  | E4 | 0.12 | 0.40 | 0.36 | 0.36 | 0.40 | 0.12 |
|  | E5 | 1.66$^2$ | 0.10 | 1.59$^2$ | 1.66$^2$ | 0.10 | 1.59$^2$ |

Table 9: Results for Group 3 SD$_D^R$ datasets showing the t-values and the superscript shows whether the null hypothesis is rejected or accepted in each case for the CIs considered (95%, 70%, 60%). Superscript '1' indicates rejection with all 3 CIs, '2' indicates rejection with 70 and 60. '3' indicates rejection with 60 %.

## D.6 HD1$_D^R$

Table 17 shows the intermediate calculations for the hypotheses stated in Section 6.4.1 of the main paper.

## D.7 HD1$_S^R$

We replicate the original experiment performed with HD1 (Section 6.1 of the main paper) but the data is roundtrip translated using Spanish as the intermediate language. Table 18 shows the intermediate calculations for this experiment.

| SAS | E. words | $RG_nRG_{em}$ | $RG_nRG_{ef}$ | $RG_nRG_{am}$ | $RG_nRG_{af}$ | $RG_{em}RG_{ef}$ | $RG_{em}RG_{am}$ | $RG_{em}RG_{af}$ | $RG_{ef}RG_{am}$ | $RG_{ef}RG_{af}$ | $RG_{am}RG_{af}$ |
|---|---|---|---|---|---|---|---|---|---|---|---|
| $S_b$ | E1 | 0 | H[1] | 0 | H[1] | H[1] | 0 | H[1] | H[1] | 0 | H[1] |
|  | E2 | 0 | H[1] | 0 | H[1] | H[1] | 0 | H[1] | H[1] | 0 | H[1] |
|  | E3 | 0 | H[1] | 0 | H[1] | H[1] | 0 | H[1] | H[1] | 0 | H[1] |
|  | E4 | 0 | H[1] | 0 | H[1] | H[1] | 0 | H[1] | H[1] | 0 | H[1] |
|  | E5 | 0 | H[1] | 0 | H[1] | H[1] | 0 | H[1] | H[1] | 0 | H[1] |
| $S_r$ | E1 | 0.41 | 0.61 | 1.04 | 0.02 | 0.34 | 0.84 | 0.40 | 0.42 | 0.60 | 1.05 |
|  | E2 | 0.84 | 0.02 | 0.11 | 0.78 | 0.89 | 0.80 | 0.07 | 0.14 | 0.82 | 0.75 |
|  | E3 | 1.94[2] | 4.38[1] | 2.65[1] | 0.09 | 1.47[3] | 0.84 | 1.54[2] | 0.27 | 3.18[1] | 2.21[2] |
|  | E4 | 1.07 | 0.95 | 0.59 | 1.25 | 0.17 | 0.54 | 0.19 | 0.38 | 0.37 | 0.73 |
|  | E5 | 0.73 | 0.17 | 0.41 | 0.47 | 0.74 | 0.97 | 1.01 | 0.18 | 0.23 | 0.05 |
| $S_t$ | E1 | 0 | 0 | 0 | 0 | 0 | 0 | 0 | 0 | 0 | 0 |
|  | E2 | 0 | 0 | 0 | 0 | 0 | 0 | 0 | 0 | 0 | 0 |
|  | E3 | 0 | 0 | 0 | 0 | 0 | 0 | 0 | 0 | 0 | 0 |
|  | E4 | 0.13 | 0.13 | 0.13 | 0.13 | 0 | 0 | 0 | 0 | 0 | 0 |
|  | E5 | 0 | 0.12 | 0.12 | 0.12 | 0.10 | 0.10 | 0.10 | 0 | 0 | 0 |
| $S_d$ | E1 | 0 | 0 | 0 | 0 | 0 | 0 | 0 | 0 | 0 | 0 |
|  | E2 | 0 | 0 | 0 | 0 | 0 | 0 | 0 | 0 | 0 | 0 |
|  | E3 | 0 | 0 | 0 | 0 | 0 | 0 | 0 | 0 | 0 | 0 |
|  | E4 | 0 | 0 | 0 | 0 | 0 | 0 | 0 | 0 | 0 | 0 |
|  | E5 | 0 | 0 | 0 | 0 | 0 | 0 | 0 | 0 | 0 | 0 |
| $S_g$ | E1 | 1.98[2] | 0 | 1.98[2] | 1.98[2] | 1.02 | 0 | 1.02 | 1.02 | 1.02 | 0 |
|  | E2 | 1.06 | 1.06 | 1.06 | 1.06 | 0 | 0 | 0 | 0 | 0 | 0 |
|  | E3 | 0.65 | 0.21 | 0.21 | 0.21 | 0.65 | 0 | 0 | 0.65 | 0.65 | 0 |
|  | E4 | 0.18 | 0.70 | 0 | 0.18 | 0.69 | 0.14 | 0 | 0.53 | 0.69 | 0.14 |
|  | E5 | 1.25 | 1.30 | 1.30 | 1.15 | 0 | 0 | 0.15 | 0 | 0.15 | 0.15 |

Table 10: Results for Group 3 $SD_D^R$ composite case datasets (when output sentiment values are discretized) showing the t-values and the superscript shows whether the null hypothesis is rejected or accepted in each case for the CIs considered (95%, 70%, 60%). Superscript '1' indicates rejection with all 3 CIs, '2' indicates rejection with 70 and 60. '3' indicates rejection with 60 %.

## D.8 HD2$_D^R$

Table 19 shows the intermediate calculations for the experiment described in Section 6.4.2 of the supplementary.

## D.9 HD2$_S^R$

We replicate the original experiment performed with HD2 (Section 6.2 of the main paper) but the data is roundtrip translated using Spanish as the intermediate language. Table 20 shows the intermediate calculations for this experiment.

| SAS | E.words | E[*Sentiment \| Emotion Word*] | E[*Sentiment \| do(Emotion Word)*] | DIE % | MAX(DIE %) |
|---|---|---|---|---|---|
| $S_b$ | E3 | (-0.16,-0.50) | (-0.08,-0.08) | (50,84) | 84 |
|  | E4 | (-0.20,-0.55) | (-0.10,0.03) | (50,105.4) | 105.4 |
|  | E5 | (0.11,-0.60) | (0.03,-0.11) | (72.72,74.24) | 74.24 |
| $S_r$ | E3 | (0.09,0.18) | (0.12,0.26) | (33.33,44.44) | 44.44 |
|  | E4 | (0.20,0.16) | (0.20,0.12) | (0,25) | 25 |
|  | E5 | (0.05,-0.30) | (0.13,-0.30) | (160,0) | 160 |
| $S_t$ | E3 | (0,0.8) | (0,0.8) | (0,0) | 0 |
|  | E4 | (-0.12,0.8) | (-0.12,0.8) | (0,0) | 0 |
|  | E5 | (0,0.8) | (0,0.8) | (0,0) | 0 |
| $S_d$ | E3 | (-1,1) | (-0.23,0.77) | (77,33) | 77 |
|  | E4 | (-1,1) | (-0.28,0.78) | (72,22) | 72 |
|  | E5 | (-1,1) | (-0.20,0.79) | (80,21) | 80 |
| $S_g$ | E3 | (-0.61,-0.14) | (-0.58,-0.16) | (4.91,14.28) | 14.28 |
|  | E4 | (-0.62,-0.01) | (-0.58,-0.02) | (6.45,100) | 100 |
|  | E5 | (-0.55,-0.13) | (-0.53,-0.14) | (3.63,7.69) | 7.69 |

Table 11: E[Sentiment | Emotion Word] and E[Sentiment | do(Emotion Word)] values for Group 4 $SD_D^R$ datasets created using round-tripped data when Danish is used as an intermediate language ($SD_D^R$) and the DIE % when emotion word sets, E3, E4 and E5 are considered. We then compute the MAX() from the DIE %

| SAS | E. words | $G_mG_n$ | $G_mG_f$ | $G_fG_n$ |
|---|---|---|---|---|
| $S_b$ | E1 | 0 | H[1] | H[1] |
|  | E2 | 0 | H[1] | H[1] |
|  | E3 | 0 | H[1] | H[1] |
|  | E4 | 0 | H[1] | H[1] |
|  | E5 | 0 | H[1] | H[1] |
| $S_r$ | E1 | 1.59 [2] | 0.62 | 1.20 |
|  | E2 | 0.31 | 0.33 | 0.65 |
|  | E3 | 0.53 | 0.01 | 0.50 |
|  | E4 | 0.83 | 1.22 | 0.44 |
|  | E5 | 0.25 | 0.45 | 0.70 |
| $S_t$ | E1 | 1.52 [2] | 0 | 1.52 [2] |
|  | E2 | 0 | 0 | 0 |
|  | E3 | 0.38 | 0 | 0.38 |
|  | E4 | 0.59 | 0.04 | 0.54 |
|  | E5 | 0.27 | 0.18 | 0.47 |
| $S_d$ | E1 | 0 | 0 | 0 |
|  | E2 | 0 | 0 | 0 |
|  | E3 | 0 | 0 | 0 |
|  | E4 | 0 | 0 | 0 |
|  | E5 | 0 | 0 | 0 |
| $S_g$ | E1 | 0.67 | 0.45 | 0.29 |
|  | E2 | 1.67 [2] | 1.39 [3] | 0.47 |
|  | E3 | 1.59 [2] | 1.18 | 0.54 |
|  | E4 | 1.46 [2] | 0.83 | 0.89 |
|  | E5 | 1.96 [2] | 1.28 | 0.97 |

Table 12: Results for Group-1 $SD_S^R$ datasets showing the t-values and the superscript shows whether the null hypothesis is rejected or accepted in each case for the CIs considered (95%, 70%, 60%). Superscript '1' indicates rejection with all 3 CIs, '2' indicates rejection with 70 and 60. '3' indicates rejection with 60 %.

### D.10 $S_h$ on Round-trip translated data

The t-values for Groups 1,3 of $SD_D^R$, $SD_S^R$ and $HD1_D^R$, $HD1_S^R$, $HD2_D^R$, $HD2_S^R$ are all 0s (as the means of the distributions (Sentiment | Gender), (Sentiment | Race) and (Sentiment | RG) are equal for all classes (hence, t-values turned out to be 0s).

T-values for Groups 2,4 of $SD_D^R$ and $SD_S^R$ are shown in Tables 21, 22, 23 and 24 respectively.

## E   Research Questions and Interpretations

In this section, we draw insights from the experimental results and give an interpretation of the observed results. The final conclusions which answer these research questions are in the main paper (Section 6.5).
**RQ1: For prototypical SAS approaches, how does sentiment rating on human-generated data compare with synthetic data?**

| SAS | E.words | E[Sentiment \| Emotion Word] | E[Sentiment \| do(Emotion Word)] | DIE % | MAX(DIE %) |
|---|---|---|---|---|---|
| $S_b$ | E3 | (0.23,-1) | (-0.08,-0.24) | (134.7, 76) | 76 |
| | E4 | (0.40,-0.85) | (0.02,-0.16) | (95,81.17) | 95 |
| | E5 | (0.14,-1) | (-0.04,-0.28) | (128.5, 72) | 128.5* |
| $S_r$ | E3 | (0.18,0.20) | (0.16,0.22) | (11.11, 10) | 11.11 |
| | E4 | (-0.09,0.09) | (0.02,0) | (122.22, 100) | 122.22* |
| | E5 | (-0.27,-0.15) | (-0.29,-0.09) | (7.4, 40) | 40 |
| $S_t$ | E3 | (-0.07,0.80) | (-0.05,0.78) | (28.57, 2.5) | 28.57* |
| | E4 | (-0.07,0.80) | (-0.05,0.79) | (28.57, 1.25) | 28.57* |
| | E5 | (-0.08,0.80) | (-0.06,0.78) | (25, 2.5) | 25 |
| $S_d$ | E3 | (-1,1) | (-0.26,0.76) | (74, 24) | 74 |
| | E4 | (-1,1) | (-0.26,0.79) | (74, 21) | 74 |
| | E5 | (-1,1) | (-0.22,0.79) | (78, 21) | 78* |
| $S_g$ | E3 | (-0.38,0.1) | (-0.38,0.1) | (0,0) | 0 |
| | E4 | (-0.44,-0.06) | (-0.45,0.01) | (2.27, 116.66) | 116.66* |
| | E5 | (-0.38,-0.08) | (-0.42,-0.02) | (10.52, 75) | 75 |

Table 13: E[Sentiment | Emotion Word] and E[Sentiment | do(Emotion Word)] values for Group 2 $SD_S^R$ datasets and the DIE % when emotion word sets, E3, E4 and E5 are considered. We then compute the MAX() from the DIE %. Superscript '*' in MAX(DIE %) denotes the MAX(MAX(DIE %))

**Observations:**
**Human-generated Data (HD):** In HD1 and HD2, there is no confounding effect. To get the raw scores, we only computed WRS. From HD1 results in Table 3 of the main paper, there is no bias in the output sentiment of chatbot responses but $S_g$ showed the highest amount of bias in user responses and $S_d$ was second most biased system. From HD2 results in Table 3 of the main paper, all the three deployed SASs i.e., $S_d, S_g, S_t$ showed bias in both chatbot and user conversations. $S_g$ showed highest amount of bias in chatbot responses but $S_t$ and $S_d$ showed highest amount of bias in user responses.
**Synthetic Data (SD):** From the SD results in Table 2 of the main paper, $S_d$ showed high amount of confounding bias compared to other systems (observed from results of Groups 2 and 4), second highest being $S_d$ and $S_g$ was the third highest. In the absence of confounders, only $S_g$ showed some bias when tested on Group-1 datasets. These conclusions can be drawn from the same table.
**RQ2: How does rating of prototypical SAS approaches compare with human annotation?**
**Observations:**
**Human-generated Data (HD):** Table 3 of the main paper shows the partial order (with raw scores computed using WRS) and complete order (with final ratings) computed for all the SASs along with '$S_h$' (human annotated sentiment analyzer). This system showed no bias. The WRS was zero for both HD1 and HD2 datasets.
**Synthetic Data (SD):** Table 2 of the main paper shows the partial order (with raw scores computed using DIE or WRS) and complete order (with final ratings) which were computed in [7]. For Groups 1 and 3, $S_h$ did not show any statistical bias (bias when there is no confounding effect). For Groups 2 and 4, the computed DIE % was 100 (confounding bias in the presence of confounder).
**RQ3: How does rating of prototypical SAS approaches get impacted when text is round-tripped between English and other languages?**

16

| SAS | E. words | $G_mG_n$ | $G_mG_f$ | $G_fG_n$ | $R_eR_n$ | $R_eR_a$ | $R_aR_n$ |
|---|---|---|---|---|---|---|---|
| $S_b$ | E1 | 0 | H[1] | H[1] | 2.64[1] | 0 | 2.64[1] |
| | E2 | 0 | H[1] | H[1] | 2.64[1] | 0 | 2.64[1] |
| | E3 | 0 | H[1] | H[1] | 3.87[1] | 0 | 3.87[1] |
| | E4 | 0 | H[1] | H[1] | 4.80[1] | 0 | 4.80[1] |
| | E5 | 0 | H[1] | H[1] | 4.80[1] | 0 | 4.80[1] |
| $S_r$ | E1 | 0.47 | 1.61 [2] | 1 | 0.21 | 1.04 | 0.74 |
| | E2 | 1.21 | 1.28 | 0.02 | 0.19 | 0.90 | 1.02 |
| | E3 | 0.75 | 1.68 [2] | 0.76 | 0.13 | 0.33 | 0.17 |
| | E4 | 1.35 | 1.09 | 0.28 | 1.16 [3] | 0.60 | 0.50 |
| | E5 | 0.57 | 0.19 | 0.38 | 0.29 | 0.37 | 0.65 |
| $S_t$ | E1 | 1 | 0 | 1 | 1.52 [2] | 1.52 [2] | 0 |
| | E2 | 0 | 0 | 0 | 0 | 0 | 0 |
| | E3 | 0.2 | 0 | 0.2 | 0.4 | 0.4 | 0 |
| | E4 | 0.31 | 0.04 | 0.36 | 0.59 | 0.54 | 0.06 |
| | E5 | 0.14 | 0.14 | 0.28 | 0.27 | 0.13 | 0.15 |
| $S_d$ | E1 | 0 | 0 | 0 | 0 | 0 | 0 |
| | E2 | 0 | 0 | 0 | 0 | 0 | 0 |
| | E3 | 0 | 0 | 0 | 0 | 0 | 0 |
| | E4 | 0 | 0 | 0 | 0 | 0 | 0 |
| | E5 | 0 | 0 | 0 | 0 | 0 | 0 |
| $S_g$ | E1 | 2.82[1] | 1.55[2] | 0 | 0.96 | 0 | 0.96 |
| | E2 | 1.91[2] | 0.63 | 2.57 [1] | 2.57 [1] | 0.63 | 1.91 [2] |
| | E3 | 0.13 | 1.14 | 1.56[2] | 1 | 0.30 | 0.66 |
| | E4 | 0.34 | 1.05 | 1.62[2] | 1.17 | 0.34 | 0.79 |
| | E5 | 2.26[2] | 0.65 | 2.90 [1] | 2.90 [1] | 0.65 | 2.26 [2] |

Table 14: Results for Group 3 $SD_S^R$ datasets showing the t-values and the superscript shows whether the null hypothesis is rejected or accepted in each case for the CIs considered (95%, 70%, 60%). Superscript '1' indicates rejection with all 3 CIs, '2' indicates rejection with 70 and 60. '3' indicates rejection with 60 %.

| SAS | E. words | $RG_nRG_{em}$ | $RG_nRG_{ef}$ | $RG_nRG_{am}$ | $RG_nRG_{af}$ | $RG_{em}RG_{ef}$ | $RG_{em}RG_{am}$ | $RG_{em}RG_{af}$ | $RG_{ef}RG_{am}$ | $RG_{ef}RG_{af}$ | $RG_{am}RG_{af}$ |
|---|---|---|---|---|---|---|---|---|---|---|---|
| $S_b$ | E1 | 0 | H[1] | 0 | H[1] | H[1] | 0 | H[1] | H[1] | 0 | H[1] |
|  | E2 | 0 | H[1] | 0 | H[1] | H[1] | 0 | H[1] | H[1] | 0 | H[1] |
|  | E3 | 0 | H[1] | 0 | H[1] | H[1] | 0 | H[1] | H[1] | 0 | H[1] |
|  | E4 | 0 | H[1] | 0 | H[1] | H[1] | 0 | H[1] | H[1] | 0 | H[1] |
|  | E5 | 0 | H[1] | 0 | H[1] | H[1] | 0 | H[1] | H[1] | 0 | H[1] |
| $S_r$ | E1 | 0.02 | 0.40 | 0.87 | 2.07 [2] | 0.42 | 0.89 | 2.02 [2] | 0.50 | 2.49 [2] | 2.88 [2] |
|  | E2 | 0.14 | 0.43 | 3.16 [1] | 0.46 | 0.53 | 3 [1] | 0.29 | 1.85 [2] | 0.80 | 3.61 [1] |
|  | E3 | 0.52 | 0.25 | 0.72 | 1.08 | 0.68 | 0.18 | 1.51 [3] | 0.85 | 0.63 | 1.72 [2] |
|  | E4 | 2.62 [1] | 0.48 | 0.04 | 0.87 | 2.80 [1] | 2.08 [2] | 1.30 | 0.34 | 1.20 | 0.76 |
|  | E5 | 0.56 | 0.07 | 0.36 | 0.71 | 0.56 | 0.17 | 0.12 | 0.38 | 0.69 | 0.29 |
| $S_t$ | E1 | 1 | 1 | 0 | 0 | 0 | 1 | 1 | 1 | 1 | 0 |
|  | E2 | 0 | 0 | 0 | 0 | 0 | 0 | 0 | 0 | 0 | 0 |
|  | E3 | 0.29 | 0.29 | 0 | 0 | 0 | 0.26 | 0.26 | 0.26 | 0.26 | 0 |
|  | E4 | 0.49 | 0.42 | 0.03 | 0.13 | 0.06 | 0.47 | 0.33 | 0.41 | 0.27 | 0.14 |
|  | E5 | 0.25 | 0.18 | 0.03 | 0.28 | 0.06 | 0.25 | 0 | 0.19 | 0.07 | 0.27 |
| $S_d$ | E1 | 0 | 0 | 0 | 0 | 0 | 0 | 0 | 0 | 0 | 0 |
|  | E2 | 0 | 0 | 0 | 0 | 0 | 0 | 0 | 0 | 0 | 0 |
|  | E3 | 0 | 0 | 0 | 0 | 0 | 0 | 0 | 0 | 0 | 0 |
|  | E4 | 0 | 0 | 0 | 0 | 0 | 0 | 0 | 0 | 0 | 0 |
|  | E5 | 0 | 0 | 0 | 0 | 0 | 0 | 0 | 0 | 0 | 0 |
| $S_g$ | E1 | 1.97 [2] | 0 | 1.97 [2] | 0 | 1.02 | 0 | 1.02 | 1.02 | 0 | 1.02 |
|  | E2 | 1.99 [2] | 1.99 [2] | 1.05 | 1.99 [2] | 0 | 0.87 | 0 | 0.87 | 0 | 0.87 |
|  | E3 | 0.35 | 1.13 | 0.21 | 1.13 | 0.54 | 0.43 | 0.54 | 1.05 | 0 | 1.05 |
|  | E4 | 0.53 | 1.18 | 0.09 | 1.18 | 0.46 | 0.50 | 0.46 | 1.03 | 0 | 1.03 |
|  | E5 | 2.18 [2] | 2.18 [2] | 1.24 | 2.18 [2] | 0 | 0.96 | 0 | 0.96 | 0 | 0.96 |

Table 15: Results for Group 3 $SD_S^R$ composite case datasets showing the t-values and the superscript shows whether the null hypothesis is rejected or accepted in each case for the CIs considered (95%, 70%, 60%). Superscript '1' indicates rejection with all 3 CIs, '2' indicates rejection with 70 and 60. '3' indicates rejection with 60 %.

**Observations:**
**Human-generated Data (HD):** High WRS indicates high bias. There is no change in WRS for any of the SASs when $HD1_D^R$ and $HD1_S^R$ chatbot conversations were used. There is a slight increase in bias when the SASs were tested on $HD1_S^R$. When SASs were tested on chatbot conversations of $HD2_D^R$ and $HD2_S^R$, the WRS of $S_g, S_t, S_d$ either decreased or remained same. When user conversations of the same dataset were used, only the WRS of $S_t$ decreased. This can be observed from Table 3 of the main paper.
**Synthetic Data (SD):** High DIE % indicates high confounding bias. The WRS (in the absence of confounder) and DIE % (in the presence of confounder) of $S_g$ increased when SD is round-trip translated. DIE % of $S_t$ decreased when SD is round-trip translated i.e., the confounding bias decreased after round-trip translation.

| SAS | E.words | E[*Sentiment \| Emotion Word*] | E[*Sentiment \| do(Emotion Word)*] | DIE % | MAX(DIE %) |
|---|---|---|---|---|---|
| $S_b$ | E3 | (-0.16,-0.50) | (-0.08,-0.08) | (50,84) | 84 |
|  | E4 | (-0.20,-0.55) | (-0.10,0.03) | (50,105.4) | 105.4* |
|  | E5 | (0.11,-0.60) | (0.03,-0.11) | (72.72,74.24) | 74.24 |
| $S_r$ | E3 | (0.11,0.23) | (0.12,0.26) | (9.09, 13.04) | 13.04 |
|  | E4 | (0.08,0.27) | (0.13,0.18) | (62.5, 33.33) | 62.5* |
|  | E5 | (-0.09,-0.23) | (-0.07,-0.29) | (22.22, 26.08) | 26.08 |
| $S_t$ | E3 | (-0.04,0.8) | (-0.04,0.8) | (0,0) | 0 |
|  | E4 | (-0.11,0.8) | (-0.11,0.8) | (0,0) | 0 |
|  | E5 | (0,0.8) | (0,0.8) | (0,0) | 0 |
| $S_d$ | E3 | (-1,1) | (-0.23,0.77) | (77,33) | 77 |
|  | E4 | (-1,1) | (-0.28,0.78) | (72,22) | 72 |
|  | E5 | (-1,1) | (-0.20,0.79) | (80,21) | 80* |
| $S_g$ | E3 | (-0.50,0.11) | (-0.50,0.15) | (0, 36.36) | 36.36* |
|  | E4 | (-0.53,0.11) | (-0.52,0.14) | (1.88, 27.27) | 27.27 |
|  | E5 | (-0.40,0) | (-0.44,0.02) | (10, X) | 10, $X^*$ |

Table 16: E[Sentiment | Emotion Word] and E[Sentiment | do(Emotion Word)] values for Group 4 $SD_S^R$ datasets and the DIE % when emotion word sets, E3, E4 and E5 are considered. We then compute the MAX() from the DIE %. Superscript '*' in MAX(DIE %) denotes the MAX(MAX(DIE %))

| Compared Distribution | SAS | $G_m G_f$ |
|---|---|---|
| (Sentiment of user responses \| Gender) | $S_b$ | H[1] |
|  | $S_r$ | 0.54 |
|  | $S_t$ | 1.28 |
|  | $S_d$ | 1.33[3] |
|  | $S_g$ | 3.47[1] |
| (Sentiment of chatbot responses \| Gender) | $S_b$ | H[1] |
|  | $S_r$ | 1.53[2] |
|  | $S_t$ | 0.37 |
|  | $S_d$ | 1.04 |
|  | $S_g$ | 1.20 |

Table 17: Results showing t-values when (HD1$_D^R$) is used. The superscript shows whether the null hypothesis is rejected or accepted in each case for the CIs considered (95%, 70%, 60%) when the distributions, (Sentiment of user responses | Gender) and (Sentiment of chatbot responses | Gender) are compared across different genders. Superscript '1' indicates rejection with all 3 CIs, and '2' indicates rejection with 70 % and 60 %, '3' indicates rejection with 60 %.

| Compared Distribution | SAS | $G_mG_f$ |
|---|---|---|
| (Sentiment of user responses \| Gender) | $S_b$ | H[1] |
| | $S_r$ | 1 |
| | $S_t$ | 0 |
| | $S_d$ | 1.33[3] |
| | $S_g$ | 3.50[1] |
| (Sentiment of chatbot responses \| Gender) | $S_b$ | H[1] |
| | $S_r$ | 2.15[2] |
| | $S_t$ | 0.15 |
| | $S_d$ | 1.04 |
| | $S_g$ | 0.25 |

Table 18: Results showing t-values when (HD1$_S^R$) is used. The superscript shows whether the null hypothesis is rejected or accepted in each case for the CIs considered (95%, 70%, 60%) when the distributions, (Sentiment of user responses | Gender) and (Sentiment of chatbot responses | Gender) are compared across different genders. Superscript '1' indicates rejection with all 3 CIs, and '2' indicates rejection with 70 % and 60 %,'3' indicates rejection with 60 %.

| Compared Distribution | SAS | $G_mG_n$ | $G_mG_f$ | $G_fG_n$ |
|---|---|---|---|---|
| (Sentiment of user responses \| Gender) | $S_b$ | 0 | H[1] | H[1] |
| | $S_r$ | 0.45 | 0.65 | 0.20 |
| | $S_t$ | 0.80 | 2.74[1] | 3.60[1] |
| | $S_d$ | 4.14[1] | 1.83[2] | 6.76[1] |
| | $S_g$ | 6.50[1] | 4.65[1] | 0.79 |
| (Sentiment of chatbot responses \| Gender) | $S_b$ | 0 | H[1] | H[1] |
| | $S_r$ | 0.56 | 0.45 | 0.13 |
| | $S_t$ | 1.11 | 1.52[2] | 2.88[1] |
| | $S_d$ | 1.24 | 0.50 | 1.91[2] |
| | $S_g$ | 0.74 | 1.91[2] | 1.38[3] |

Table 19: Results showing t-values when (HD2$_D^R$) is used. The superscript shows whether the null hypothesis is rejected or accepted in each case for the CIs considered (95%, 70%, 60%) when the distributions, (Sentiment of user responses | Gender) and (Sentiment of chatbot responses | Gender) are compared across different genders. Superscript '1' indicates rejection with all 3 CIs, and '2' indicates rejection with 70 % and 60 %, '3' indicates rejection with 60 %.

| Compared Distribution | SAS | $G_mG_n$ | $G_mG_f$ | $G_fG_n$ |
|---|---|---|---|---|
| (Sentiment of user responses \| Gender) | $S_b$ | 0 | H[1] | H[1] |
| | $S_r$ | 1.31[3] | 0.30 | 1.70[2] |
| | $S_t$ | 2.62[1] | 2.94[1] | 0.82 |
| | $S_d$ | 3.08[1] | 1.33[3] | 1.74[2] |
| | $S_g$ | 1.44[2] | 2.33[1] | 3.24[1] |
| (Sentiment of chatbot responses \| Gender) | $S_b$ | 0 | H[1] | H[1] |
| | $S_r$ | 0.67 | 0.65 | 1.50[2] |
| | $S_t$ | 0.54 | 1.46[2] | 2.20[2] |
| | $S_d$ | 0.46 | 1.04 | 0.09 |
| | $S_g$ | 1.04 | 3.41[1] | 4.86[1] |

Table 20: Results showing t-values when (HD2$_S^R$) is used. The superscript shows whether the null hypothesis is rejected or accepted in each case for the CIs considered (95%, 70%, 60%) when the distributions, (Sentiment of user responses | Gender) and (Sentiment of chatbot responses | Gender) are compared across different genders. Superscript '1' indicates rejection with all 3 CIs, and '2' indicates rejection with 70 % and 60 %, '3' indicates rejection with 60 %.

| E.words | E[*Sentiment \| Emotion Word*] | E[*Sentiment \| do(Emotion Word)*] | DIE % | MAX(DIE %) |
|---|---|---|---|---|
| E3 | (-1,1) | (-0.26,0.78) | (74,22) | 74 |
| E4 | (-1,1) | (-0.17,0.88) | (83,12) | 83 |
| E5 | (-1,1) | (-0.26,0.77) | (74,23) | 74 |

Table 21: E[Sentiment | Emotion Word] and E[Sentiment | do(Emotion Word)] values for the system $S_h$ on Group-2 datasets of SD$_D^R$ and the DIE % when emotion word sets, E3, E4 and E5 are considered. We consider the worst possible case using MAX(). %

| E.words | E[*Sentiment \| Emotion Word*] | E[*Sentiment \| do(Emotion Word)*] | DIE % | MAX(DIE %) |
|---|---|---|---|---|
| E3 | (-1,1) | (-0.23,0.77) | (77,23) | 77 |
| E4 | (-1,1) | (-0.28,0.78) | (72,22) | 72 |
| E5 | (-1,1) | (-0.2,0.79) | (80,21) | 80 |

Table 22: E[Sentiment | Emotion Word] and E[Sentiment | do(Emotion Word)] values for the system $S_h$ on Group-4 datasets of SD$_D^R$ and the DIE % when emotion word sets, E3, E4 and E5 are considered. We consider the worst possible case using MAX(). %

| E.words | E[*Sentiment \| Emotion Word*] | E[*Sentiment \| do(Emotion Word)*] | DIE % | MAX(DIE %) |
|---|---|---|---|---|
| E3 | (-1,1) | (-0.26,0.77) | (74,23) | 74 |
| E4 | (-1,1) | (-0.26,0.79) | (74,21) | 74 |
| E5 | (-1,1) | (-0.23,0.80) | (77,20) | 77 |

Table 23: E[Sentiment | Emotion Word] and E[Sentiment | do(Emotion Word)] values for the system $S_h$ on Group-2 datasets of SD$_S^R$ and the DIE % when emotion word sets, E3, E4 and E5 are considered. We consider the worst possible case using MAX(). %

| E.words | E[*Sentiment \| Emotion Word*] | E[*Sentiment \| do(Emotion Word)*] | DIE % | MAX(DIE %) |
|---|---|---|---|---|
| E3 | (-1,1) | (-0.23,0.77) | (77,23) | 77 |
| E4 | (-1,1) | (-0.28,0.78) | (72,22) | 72 |
| E5 | (-1,1) | (-0.2,0.79) | (80,21) | 80 |

Table 24: E[Sentiment | Emotion Word] and E[Sentiment | do(Emotion Word)] values for the system $S_h$ on Group-4 datasets of $SD_S^R$ and the DIE % when emotion word sets, E3, E4 and E5 are considered. We consider the worst possible case using MAX(). %

# F Instructions to Human Annotators for Annotating Sentiments for $S_h$

Each annotator received a shared drive (one per annotator) with instructions and datasets to be annotated. The annotation was done by 3 recruited volunteers who are also the authors. They include an undergraduate student, graduate student and a professor at a major university. Here are the instructions that were provided:

- There are 2 sets of datasets in this shared drive (real-world datasets and eecs-variation datasets).

- EECS datasets: These are the datasets taken from the EECS dataset. The data generation procedure was described in detail in Section 3.1 of the paper [7]. Here's the summary. EECS folder contains datasets belonging to different groups:

    - Group-1: Contains data with pronouns referring to people which serve as a proxy for their gender. In this group of datasets, gender won't affect how emotion words are distributed.
    - Group-2: Contains data with pronouns referring to people, which serve as a proxy for their gender. In this group of datasets, gender affects how emotion words are distributed.
    - Group-3: Distribution is the same as group-1, but there's an extra protected attribute: Race. The dataset has person names that serve as a proxy for both gender and race.
    - Group-4: Same as Group-2 but an extra protected attribute, race, is present.

- Real-world datasets: The provided datasets show user conversations with two different chatbots: ALLURE and Unibot.

    - ALLURE data has 1087 utterances (bot + user).
      **Attribute description:**
      C_num: Conversation number or user number.
      UB: An attribute that would indicate whether the text in the 'Text' attribute is from the user (1) or chatbot (0).
      User_gender: Gender of the user. user preferred not to answer (0), male user (1), and female user (2).
      Original: The original text stored by the chatbot.
      Enhancement: In order to include the gender information and observe the output for the experiments, we added some enhancements.
      Text: User or chatbot response.
    - Unibot data has 1281 utterances.
      **Attribute description:**
      C_num: Conversation number or user number.
      UB: An attribute that would indicate whether the text in the 'Text' attribute is from the user (1) or chatbot (0).
      User_gender: Gender of the user. user preferred not to answer (0), male user (1), and female user (2).
      Original: The original text stored by the chatbot.
      Enhancement: In order to include the gender information and observe the output for the experiments, we added some enhancements.

Text: User or chatbot response (original response + enhancement). In the ALLURE dataset, enhancements (gender information of a user) are added to input as the gender of the users who participated in the testing was known. In Unibot, gender is not known. So we added some enhancements as explained above.

If you think the sentiment of the text in the 'Text' attribute of the data is positive, please write a '1' in the new column, 'S_H.,' '-1'; if you think it is negative or '0', if you think that the text might have a neutral sentiment.

# References

[1] Elias Bareinboim and Judea Pearl. "Causal inference and the data-fusion problem". In: *Proceedings of the National Academy of Sciences* 113.27 (2016), pp. 7345–7352. DOI: [10.1073/pnas.1510507113](10.1073/pnas.1510507113). eprint: <https://www.pnas.org/doi/pdf/10.1073/pnas.1510507113>. URL: <https://www.pnas.org/doi/abs/10.1073/pnas.1510507113>.

[2] Joy Buolamwini and Timnit Gebru. "Gender Shades: Intersectional Accuracy Disparities in Commercial Gender Classification". In: *Conference on Fairness, Accountability and Transparency, FAT 2018, 23-24 February 2018, New York, NY, USA*. Ed. by Sorelle A. Friedler and Christo Wilson. Vol. 81. Proceedings of Machine Learning Research. PMLR, 2018, pp. 77–91. URL: <http://proceedings.mlr.press/v81/buolamwini18a.html>.

[3] Amir Feder et al. "Causal Inference in Natural Language Processing: Estimation, Prediction, Interpretation and Beyond". In: *CoRR* abs/2109.00725 (2021). arXiv: [2109.00725](2109.00725). URL: <https://arxiv.org/abs/2109.00725>.

[4] Joel Escudé Font and Marta R. Costa-jussà. "Equalizing Gender Biases in Neural Machine Translation with Word Embeddings Techniques". In: *CoRR* abs/1901.03116 (2019). arXiv: [1901.03116](1901.03116). URL: <http://arxiv.org/abs/1901.03116>.

[5] Md Shahriar Iqbal et al. "Unicorn: Reasoning about Configurable System Performance through the lens of Causality". In: *Proceedings of the European Conference on Computer Systems (EuroSys)*. Rennes, France, 2022.

[6] Mohammad Ali Javidian, Pooyan Jamshidi, and Marco Valtorta. "Transfer Learning for Performance Modeling of Configurable Systems: A Causal Analysis". In: *Proceedings of the First AAAI Spring Symposium Beyond Curve Fitting: Causation, Counterfactuals, and Imagination-Based AI (WHY-19)*. Stanford, CA, 2019.

[7] Kausik Lakkaraju, Biplav Srivastava, and Marco Valtorta. *Rating Sentiment Analysis Systems for Bias through a Causal Lens*. 2023. DOI: [10.48550/ARXIV.2302.02038](10.48550/ARXIV.2302.02038). URL: <https://arxiv.org/abs/2302.02038>.

[8] Chengzhi Mao et al. "Generative interventions for causal learning". In: *Proceedings of the IEEE/CVF Conference on Computer Vision and Pattern Recognition*. 2021, pp. 3947–3956.

[9] Eirini Ntoutsi et al. "Bias in Data-driven AI Systems – An Introductory Survey". In: *On Arxiv at: https://arxiv.org/abs/2001.09762*. 2020.

[10] Marcelo O. R. Prates, Pedro H. C. Avelar, and Luís C. Lamb. "Assessing Gender Bias in Machine Translation - A Case Study with Google Translate". In: *CoRR* abs/1809.02208 (2018). arXiv: [1809.02208](1809.02208). URL: <http://arxiv.org/abs/1809.02208>.

[11] Shuyuan Xu et al. *Deconfounded Causal Collaborative Filtering*. 2021. DOI: [10.48550/ARXIV.2110.07122](10.48550/ARXIV.2110.07122). URL: <https://arxiv.org/abs/2110.07122>.



\end{document}